\documentclass[10pt,twocolumn,letterpaper]{article}

\usepackage{iccv}
\usepackage{times}
\usepackage{epsfig}
\usepackage{graphicx}
\usepackage{amsmath}
\usepackage{amsfonts}
\usepackage{amssymb}
\usepackage{subcaption}

\usepackage{xcolor}
\usepackage{caption}
\usepackage{multirow}

\usepackage{amsmath,amsfonts,bm}

\def\eqref#1{equation~\ref{#1}}

\def\1{\bm{1}}

\def\vv{{\bm{v}}}

\def\mS{{\bm{S}}}

\DeclareMathAlphabet{\mathsfit}{\encodingdefault}{\sfdefault}{m}{sl}
\SetMathAlphabet{\mathsfit}{bold}{\encodingdefault}{\sfdefault}{bx}{n}

\usepackage{graphicx}
\usepackage{pifont}
\usepackage{wrapfig}
\usepackage{xspace}

\definecolor{citecolor}{rgb}{0.133, 0.752, 0.133}
\usepackage[pagebackref=true,breaklinks=true,letterpaper=true,colorlinks,bookmarks=false]{hyperref}

\makeatletter
\DeclareRobustCommand\onedot{\futurelet\@let@token\@onedot}
\def\@onedot{\ifx\@let@token.\else.\null\fi\xspace}
\def\eg{\emph{e.g}\onedot} 
\def\ie{\emph{i.e}\onedot} 

\def\wrt{w.r.t\onedot} 
\def\etal{\emph{et al}\onedot}

\makeatother

\usepackage{array}
\usepackage{colortbl}
\usepackage{diagbox}
\usepackage{booktabs}
\usepackage{boldline}
\pdfoutput=1
\usepackage{pdfpages}
\newcommand\blfootnote[1]{%
  \begingroup
  \renewcommand\thefootnote{}\footnote{#1}%
  \addtocounter{footnote}{-1}%
  \endgroup
}

\newlength\savewidth
\newcommand{\tablestyle}[2]{\setlength{\tabcolsep}{#1}\renewcommand{\arraystretch}{#2}\centering\footnotesize}

\iccvfinalcopy 

\def\iccvPaperID{8198} 

\ificcvfinal\fi

\begin{document}

\title{Rethinking Deep Contrastive Learning with Embedding Memory}

\author{Haozhi Zhang, Xun Wang, Weilin Huang$^\dagger$, Matthew R. Scott\\
Malong Technologies\\
{\tt\small \{haozhang, whuang, mscott\}@malongtech.com    bnuwangxun@gmail.com}
}

\maketitle
\ificcvfinal\thispagestyle{empty}\fi

\begin{abstract}
Pair-wise loss functions have been extensively studied and shown to continuously improve the performance of deep metric learning (DML).
However, they are primarily designed with intuition based on simple toy examples, and experimentally identifying the truly effective design is difficult in complicated, real-world cases.
In this paper, we provide a new methodology for systematically studying weighting strategies of various pair-wise loss functions, and rethink pair weighting with an embedding memory.
We delve into the weighting mechanisms by decomposing the pair-wise functions, and study positive and negative weights separately using direct weight assignment. 
This allows us to study various weighting functions deeply and systematically via weight curves, and identify a number of meaningful, comprehensive and insightful facts, which come up with our key observation on memory-based DML: it is critical to mine hard negatives and discard easy negatives which are less informative and redundant, but weighting on positive pairs is not helpful. 
This results in an efficient but surprisingly simple rule to design the weighting scheme, making it significantly different from existing mini-batch based methods which design various sophisticated loss functions to weight pairs carefully. 
Finally, we conduct extensive experiments on three large-scale visual retrieval benchmarks, and demonstrate the superiority of memory-based DML over recent mini-batch based approaches (\eg \cite{sun2020circle, zhu2020fewer, wang2019multi}), by using a simple contrastive loss with momentum-updated memory \cite{he_2019_moco}.

\blfootnote{$^\dagger$Corresponding author}
\end{abstract}

\section{Introduction}

Deep metric learning (DML) aims to learn an \emph{encoder} from raw data to an embedding space, where semantically similar instances are mapped to closer points, whereas dissimilar ones are pushed far away from each other. 
As a fundamental problem in deep leaning, DML can benefit various computer vision tasks, \eg, image retrieval \cite{wohlhart2015learning,He_2018_CVPR,irt}, face recognition \cite{Wen2016}, person re-identification \cite{Yu_2018_ECCV,in-defense}, visual tracking \cite{leal2016learning,tao2016siamese}, and zero-shot learning \cite{zhang2016zero,bucher2016improving,Yelamarthi_2018_ECCV}.

Collecting informative positive and negative pairs are of critical importance to contrastive learning.
There are mainly two directions: (1) exploring more information within a mini-batch through carefully weighting on pairs, \eg, Binomial Deviance loss \cite{binomial}, Multi-Similarity (MS) loss \cite{wang2019multi}, and Circle loss \cite{sun2020circle}; and 
(2) enriching accessible pairs during training \cite{facenet,suh2019stochastic,HTL}, \eg utilizing an embedding memory for hard negative mining \cite{wang2020cross}.
Intuitively, performing a sophisticated weighting strategy with more informative pairs should achieve better results.
However, we experimentally found that applying a simple contrastive loss with memory-based training can significantly outperform other carefully-designed loss functions (as shown in Table \ref{tab:compare_mem_loss}).
This inspired us to revisit recent pair weighting schemes in memory-based DML. 

Cross-Batch Memory (XBM) \cite{wang2020cross}, a recent memory-based DML method, was proposed by introducing an embedding memory, which allows it to collect a significantly large number of hard negatives over multiple mini-batches, and thus overcomes the fundamental limitation of mini-batch training.
%
%
With similar memory mechanism, MoCo \cite{he_2019_moco} was introduced for self-supervised representation learning, using an additional momentum updated encoder for memory updating which ensures the feature memory consistency during training.
We found that such a  momentum  encoder is of importance to reducing the ``feature drift'' problem as identified in \cite{wang2020cross}, which is critical to improving the performance of memory-based DML. 
In this work, we simply adopt MoCo in a supervised manner for DML, which is referred as s-MoCo. 
It provides a strong example for us to study the pair weighting in memory-based DML. 


General Pair Weighting (GPW) \cite{wang2019multi} was introduced  to unify various pair-based losses into a general framework, providing a powerful tool for studying  the pair weighting problem. 
Although various pair-wise loss functions have been designed in the past years, with the goal of weighting pairs more carefully. 
However, as shown in Figure \ref{fig:loss_grad}, most of these functions are highly consistent to some extend.
We extend the GPW to a direct weight assignment, which allows us to decompose the pair-wise functions, and analyze each components separately. 
This enables us to understand the underlying effect of design rules more clearly.
We delve into representative loss functions, and compare them with contrastive loss in various aspects, particularly with an embedding memory.


In this work, we rethink contrastive learning with an embedding memory, and provides a new methodology to investigate various pair-wise DML methods systematically. 
We conduct extensive experiments on three large-scale visual retrieval benchmarks. We study various weighting schemes in mini-batch training and memory-based training, and identify a number of interesting, meaningful, and insightful facts.
(1) the positive and negative weights are strongly correlated in mini-batch training, but work more independently with memory-based training;
(2) discriminatively weighting on hard positives can improve the performance with mini-batch training, but is not helpful to memory-based training;
(3) the memory module often introduces a large amount of easy negatives, which are less informative and redundant, easily making the training collapse.
(4) therefore, hard negative mining is of central importance to memory-based training, with strong robustness to the mining margin, while carefully weighting on hard negatives is again not helpful. 

These key observations result in a simple yet efficient weighting rule for memory-based DML that, \textit{hard negative mining with a rough similarity margin to discard the easy negatives is all needed}. 
Although not novel, this makes it surprisingly simple but fundamentally different from state-of-the-art mini-batch DML which always attempted to weight the pairs as accurate as possible by designing various sophisticated loss functions. We verify this rule by implementing s-MoCo, which is a contrastive loss with an embedding memory, and demonstrate  that s-MoCo can outperform state-of-the-art mini-batch methods (e.g., \cite{sun2020circle, zhu2020fewer, wang2019multi}) significantly on all three large-scale image retrieval datasets.


\section{Preliminary}
Hard sample mining and pair weighting are fundamental in DML.
In this section, we first review recent work on DML with an embedding memory, and then introduce s-MoCo as a baseline method by adapting recent MoCo \cite{he_2019_moco}.
Second, we investigate the commonality and characteristics on weighting curves of existing pair-based losses.

\begin{figure}[t]
\centering
\includegraphics[width=0.45\textwidth, trim=0 160 435 60, clip]{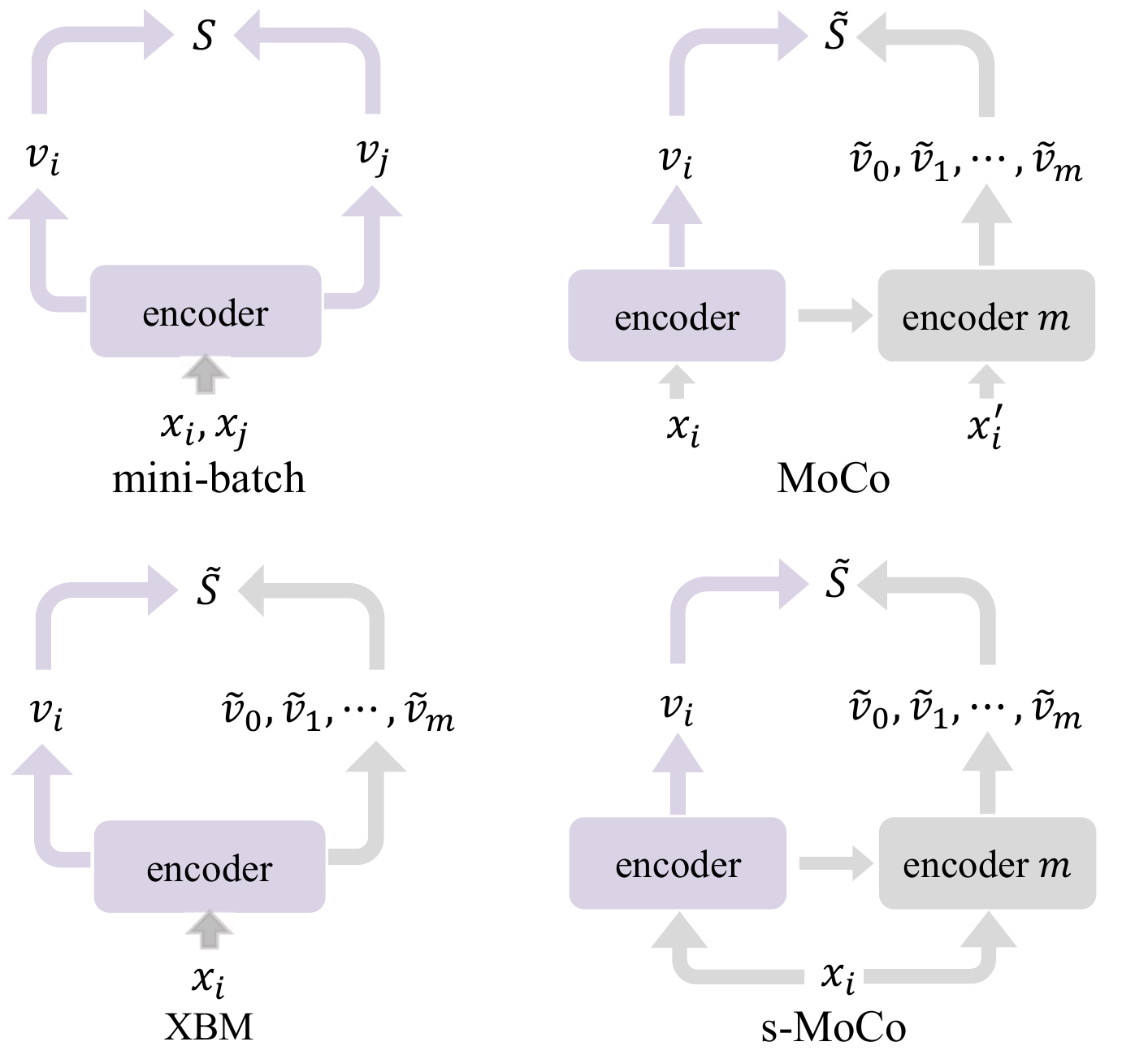}
\vspace{-0.5em}
\caption{\small \textbf{Comparison of mini-batch training, XBM, MoCo and s-MoCo.} Purple and gray arrows represent operations need back-propagation and stop-gradient respectively. $x$ is a input image and $x'$ is another data augmentation of the same image. $\widetilde{\vv}$ is a feature in the memory bank. s-MoCo is an extension of MoCo in DML.}
\label{fig:compare_moco}
\vspace{-1em}
\end{figure}

\begin{table*}[t]
\tablestyle{26pt}{2.2}
\begin{tabular}{cV{0.4}cV{0.4}c} \toprule[1pt]
Loss   & Positive Weight & Negative Weight \\ \midrule[0.3pt]
Contrastive 
& 1 
&  $\mathcal{I}_{\{S_{ij}>\lambda\}}$ \\
Binominal 
&$ \frac{ \exp\left(\alpha\left(\lambda - S_{ij} \right)\right)}{1 + \exp\left( \alpha\left(\lambda - S_{ij} \right) \right)}$ 
& $ \frac{\exp\left( \beta \left(S_{ij} - \lambda \right)\right)}{1 + \exp\left(\beta \left(S_{ij} - \lambda\right)\right)}$ 
\\
MS &
$\frac{\exp\left(\alpha \left(\lambda - S_{ij} \right)\right)}{ 1  + \sum_{k \in \mathcal{P}_i} \exp\left( \alpha \left(\lambda - S_{ik}\right)\right)}$
&
$\frac{ \exp\left(\beta \left(S_{ij} - \lambda \right)\right) }{1 + \sum_{k \in \mathcal{N}_i} \exp\left(\beta \left(S_{ik} - \lambda \right)\right)}$
\\
InfoNCE &
$-\frac{1}{\tau}\left(1 - \frac{\exp\left(S_{ij} / \tau\right)}{\exp\left(S_{ij} / \tau\right) + \sum_{n \in \mathcal{N}_i} \exp\left(S_{in} / \tau\right)}\right)$ 
& 
$\frac{1}{\tau} \sum\limits_{p \in \mathcal{P}_i} \frac{\exp\left(S_{ij} / \tau\right)}{\exp\left(S_{ip} / \tau\right) + \sum_{n \in \mathcal{N}_i} \exp\left(S_{in} / \tau\right)}$ 
\\  \bottomrule[1.1pt]
\end{tabular}%
\caption{\small \textbf{Weight assignment formulations} of four weighting strategies. $\lambda$ is the threshold of contrastive loss and Binomial loss. $\alpha$ and $\beta$ in Binomial loss or MS loss are \textbf{scale factors} for positive pairs and negative pairs respectively. $\tau$ is the temperature of InfoNCE loss.}
\label{tab:grad}%
\end{table*}%

\subsection{Hard Negative Mining with Memory}

Non-parametric memory can be viewed as an assistant module maintaining historical embeddings to overcome the limitation of mini-batch training in various computer visual tasks \cite{vinyals2016matching,xiao2017joint,wu2018improving,wu2018unsupervised}. 
It aims to cover sufficient informative samples for model training or provide global information of the whole data distribution.
In \cite{wu2018improving}, Wu \etal first applied a memory module in DML with NCA loss to optimize the similarities of positive pairs. 
On the contrary, XBM \cite{wang2020cross} was recently proposed to mine significantly more hard negative samples from an embedding memory which encodes a large number of training samples over multiple mini-batches. The embedding memory is maintained as a dynamic queue, with the current mini-batch enqueued and the oldest mini-batch dequeued.

MoCo \cite{he_2019_moco} was introduced as an outstanding method for self-supervised representation learning. It uses the same queue mechanism to maintain a feature memory, but further applies an momentum-updated memory encoder to ensure a memory consistency during training. 
Formally, the parameters of memory encoder $\theta_{M}$ is updated as below:
\vspace{-0.3em}
\begin{equation*}
\theta_{M} \leftarrow m \theta_M + (1 - m)\theta.
\vspace{-0.3em}
\end{equation*}
Note that momentum update in MoCo is applied on memory encoder, which is different from that in XBM on memory features.
This momentum update provides a strict constraint that ensures a stable update of the feature memory. We observed that such memory consistency is essentially identical to the ``feature drift" phenomenon discussed in XBM \cite{wang2020cross}, which is the key to achieve an excellent performance by XBM. In this work, we demonstrate by experiments that momentum update mechanism is able to reduce the ``feature drift" phenomenon, and better ensure the feature stability in memory-based DML, resulting in a stronger hard mining ability in both quantity and quality (Section~\ref{sec:compare_with_xbm}).


We simply adopt MoCo \cite{he_2019_moco} in a supervised manner for DML, by generating training pairs using the available category labels. 
With the category-level supervision, contrastive positive pairs can be created directly using instances from the same category, rather than using multiple augmentations of the same instance. 
We refer the supervised MoCo as s-MoCo. Similar to XBM,  s-MoCo provides a general memory-embedding framework for DML, where  most existing pair-based DML methods can be integrated, to perform hard mining in a significantly larger memory space than  the mini-batch space.  
%
Notice that, XBM can be considered as a special case of s-MoCo with $m=0$.
We compare the training frameworks of mini-batch methods, XBM, MoCo and s-MoCo in Figure~\ref{fig:compare_moco}. 

Some recent works on self-supervised learning \cite{zhu2020eqco,grill2020bootstrap} claimed that numerous hard negatives might be not necessary. However, there are strong clues indicating that the negative pairs are dispensable in DML. 
For example, as pointed out in \cite{zhu2020eqco}, the proposed EqCo did not work practically on face recognition due to its prerequisite did not met on DML,
while poor results was obtained in  \cite{wu2018improving}  by only considering the positive pairs in the memory.

\begin{figure}[t]
\begin{subfigure}{0.23\textwidth}
\centering
\includegraphics[width=1.0\linewidth, trim=10 10 10 0, clip]{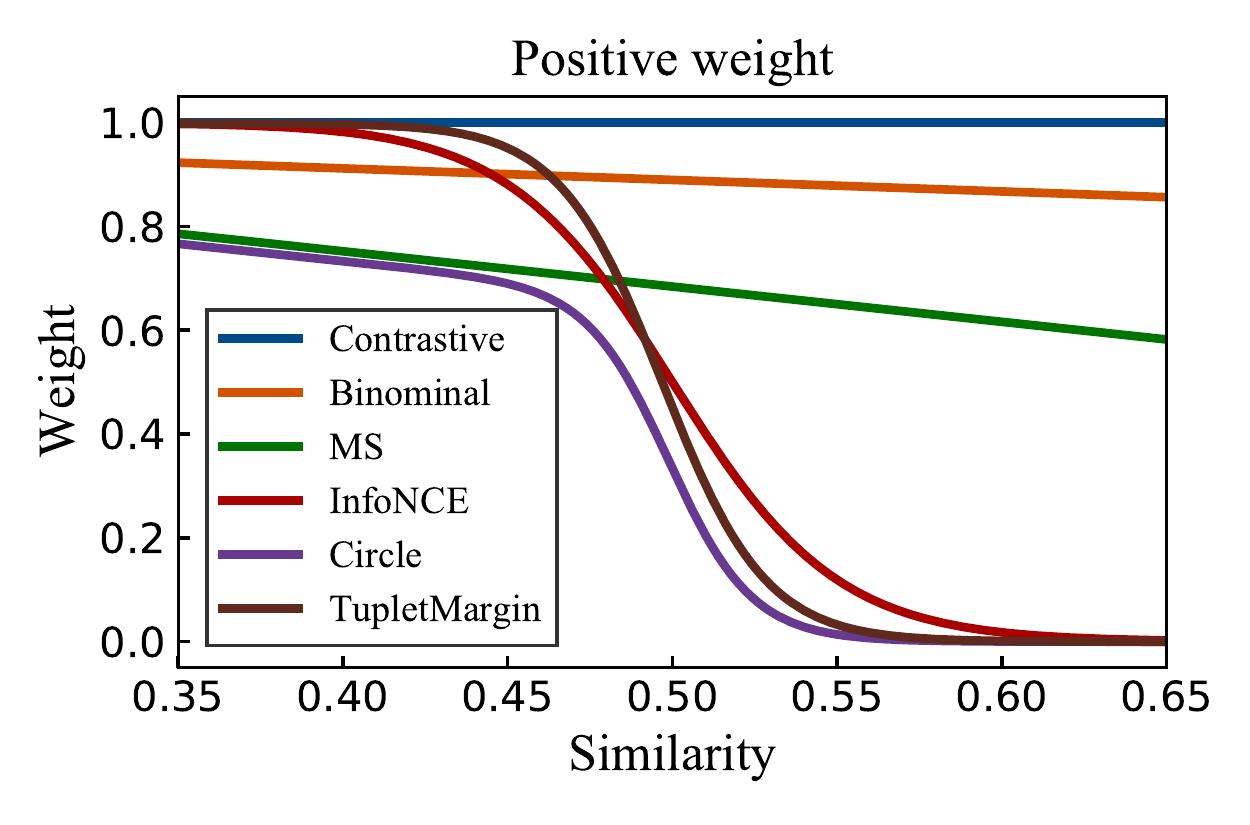}  
\end{subfigure}
\begin{subfigure}{0.23\textwidth}
\centering
\includegraphics[width=1.0\linewidth, trim=10 10 10 0, clip]{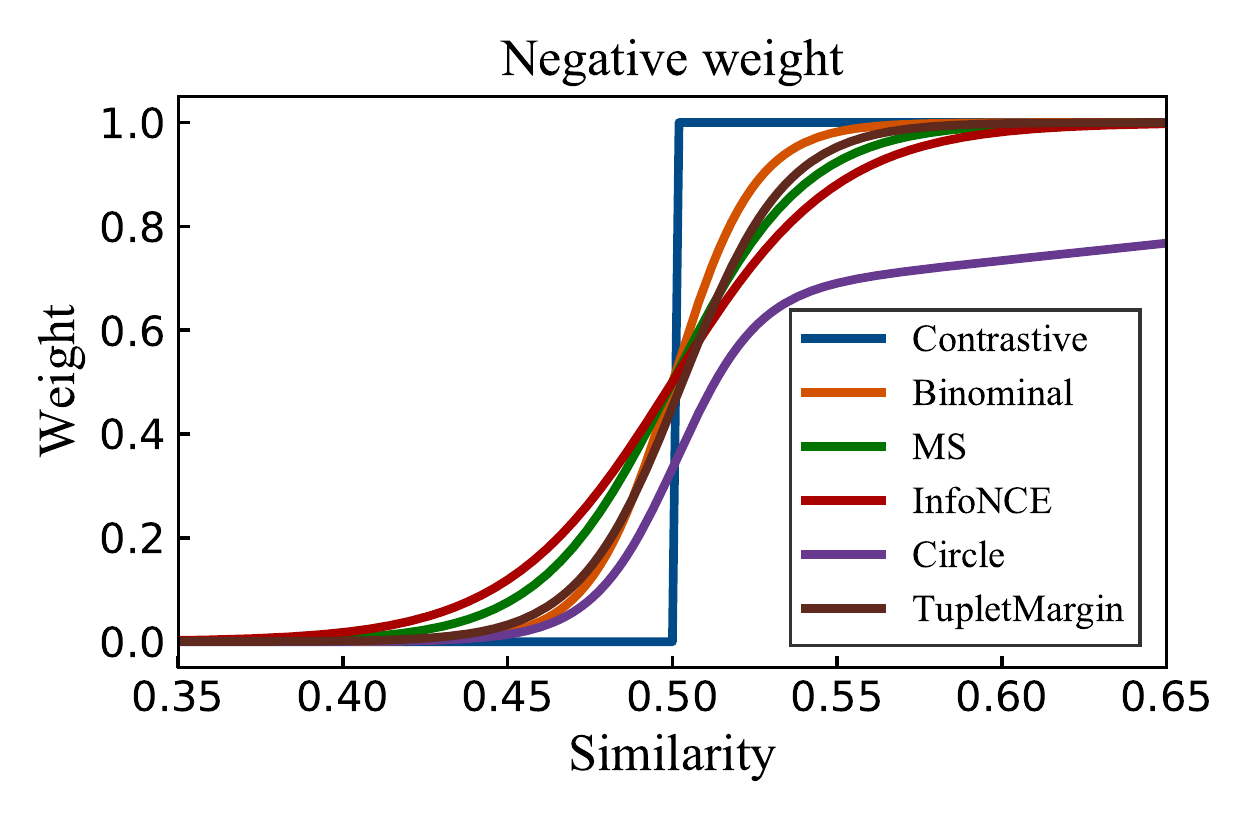}  
\end{subfigure}
\vspace{-0.5em}
\caption{\small \textbf{Positive and negative weight curves of recent state-of-the-art loss functions.} They are highly consistent to some extend and share many common characteristics\protect\footnotemark. }
\label{fig:loss_grad}
\vspace{-1em}
\end{figure}
\footnotetext{Hyper-parameters in loss functions are set as their papers. The effect of relative similarity in MS loss, Circle loss and TupletMargin loss cannot be shown in this figure.}


\subsection{Pair Weighting}
\label{sec:pair_weighting}

There are three main groups of DML approaches: pair-based \cite{contrastive,lifted-structured-loss,wang2019multi,cakir2019deep,histogram}, proxy-based \cite{proxyloss,Kim_2020_CVPR}, classification-based \cite{deng2019arcface,qian2019softtriple,zhaiclassification}.
In this work, we focus on the pair-based approaches, which are the most widely studied, with the top performance achieved.
Starting from basic Siamese networks with a contrastive loss \cite{contrastive}, a series of recent works have been dedicated to constructing more informative pairs within mini-batches \cite{facenet,n-pairs,lifted-structured-loss}. 
Although the motivations of various loss functions are diverse, they can be uniformly formulated into a general pair weighting (GPW) framework \cite{wang2019multi}:
\begin{equation*}
\small
\label{eq:weight}
 \mathcal{L}  =  \frac{1}{m} \sum_{i=1}^{m}\mathcal{L}_i = \frac{1}{m} \sum_{i=1}^{m}
\left[\sum_{j \in \mathcal{N}_i} w_{ij}  \mS_{ij} - \sum_{j \in \mathcal{P}_i} w_{ij} \mS_{ij} \right],
\end{equation*}
where $m$ is the mini-batch size, $\mathcal{P}_i$ and $\mathcal{N}_i$ are the sets of positive samples and negative samples \wrt the $i$-th anchor, and $w_{ij}$ is the weight of pair ($i$, $j$) with a  similarity of $\mS_{ij}$.
Note that $\frac{w_{ij}}{m}$ is also the partial derivative of $\mathcal{L}$ \wrt $\mS_{ij}$.
This allows us to extend GPW by directly computing the weight or gradient of each pair  from its pair similarity. 

Various weighting approaches have been developed to improve the training of deep networks by better leveraging prior knowledge \cite{binomial,wang2019multi,qian2019softtriple}. 
Toy examples were utilized to perform experimental demonstration and explain their intuitions, which may not be practical in real-world applications. 
Recent works, such as \cite{musgrave2020metric,roth2020revisiting}, compared the performance of various loss functions with fair experimental settings, in an effort to identity the key factors to training DML models.
However, they commonly performed empirical studies with off-the-shelf loss functions, and conducted experiments by varying different hyper-parameters, which however were still working in a black-box manner, to some extent.
In this work, we attempt to go beyond such direct performance comparison, and decompose loss functions by using direct weight assignment extended from GPW, which allows us to delve into pair-based DML, and deeply analyze the subtle differences among various loss functions.

With our weight assignment as described in Table \ref{tab:grad}, a pair-based loss function can be decomposed into a positive weight and a negative one, both of which are computed from the corresponding pair similarity (\eg $S_{ij}$).   
To better understand their intrinsic difference, we compare the positive and negative weights of different loss functions separately (\wrt the pair similarities) in Figure \ref{fig:loss_grad}. 
As can be found, all the loss functions follow the basic principle of DML, which encourages a high similarity for positive pairs (by weighting the low-similarity positives heavily), and a low similarity for the negative ones (by weighting the high-similarity negatives heavily).
%
Furthermore, these loss functions (except for the contrastive loss) are mainly extended from either sigmoid (\eg binominal, MS and TupletMargin loss) or softmax function (\eg InfoNCE and Circle loss).
Some recent approaches consider more information beyond self-similarity ($S_{ij}$). 
For example, MS loss \cite{wang2019multi} further computes relative similarity ($S_{ik}$) from neighboring pairs, in an effort to mine and weight the informative pairs more accurately. 
Circle loss provides a flexible optimization approach towards a more definite convergence target, while InfoNCE loss weights the positive and negative pairs jointly. 
N-pair loss \cite{n-pairs} is a special case of the InfoNCE loss with $\tau=1$.
These loss functions define different mining and weighting strategies for pair-based DML, but as shown in Figure \ref{fig:loss_grad}, their weighting curves do not have significant differences, especially on the negative weights.
Thus, we select four representative loss functions for deeply studying the pair weighting problem with an embedding memory: contrastive loss, binominal loss, MS loss and InfoNCE loss.

\section{Revisit Pair Weighting with Memory}

\begin{table}[t]
  \tablestyle{8.5pt}{1.1}
  \begin{tabular}{lV{0.4}cccc} \toprule[1pt]
  Method & Contrastive & Binominal & MS & InfoNCE \\ \midrule[0.3pt]
  mini-batch & 63.8  & 67.8  & 68.4  & 72.3 \\
  XBM  & 77.3  & 75.3  & 74.9  & 77.1 \\
  s-MoCo & \bf 79.9  & \bf 77.8  & \bf 75.6  & \bf 77.2 \\ \bottomrule[1.1pt]
  \end{tabular}%
  \vspace{-0.5em}
  \caption{\small \textbf{SOP Recall@1 with various loss functions and training schemes\protect\footnotemark.}   Contrastive loss outperform other sophisticated losses in memory-based training. s-MoCo achieves better performance than XBM.}
  \label{tab:compare_mem_loss}%
  \vspace{-1em}
  \end{table}%
  \footnotetext{Hyperparameters of binominal loss and MS loss are set as optimal stated in their papers. The hyperparameter of InfoNCE loss is optimized with grid search.}


In this section, we delve into pair-based DML methods, and study their underlying distinctions in mini-batch training and memory based training.
We adopt s-MoCo as the memory-based methods with GoogleNet \cite{inception} as the backbone network, and conduct extensive experiments on three standard benchmarks on image retrieval: SOP \cite{lifted-structured-loss}, In-Shop \cite{DeepFashion} and VehicleID \cite{liu2016deep} datasets.
More details on experimental settings are described in  Section \ref{sec:exp}.

\begin{figure}[t]
    \begin{subfigure}{0.236\textwidth}
        \centering
        \includegraphics[width=1.0\linewidth, trim=10 10 10 0, clip]{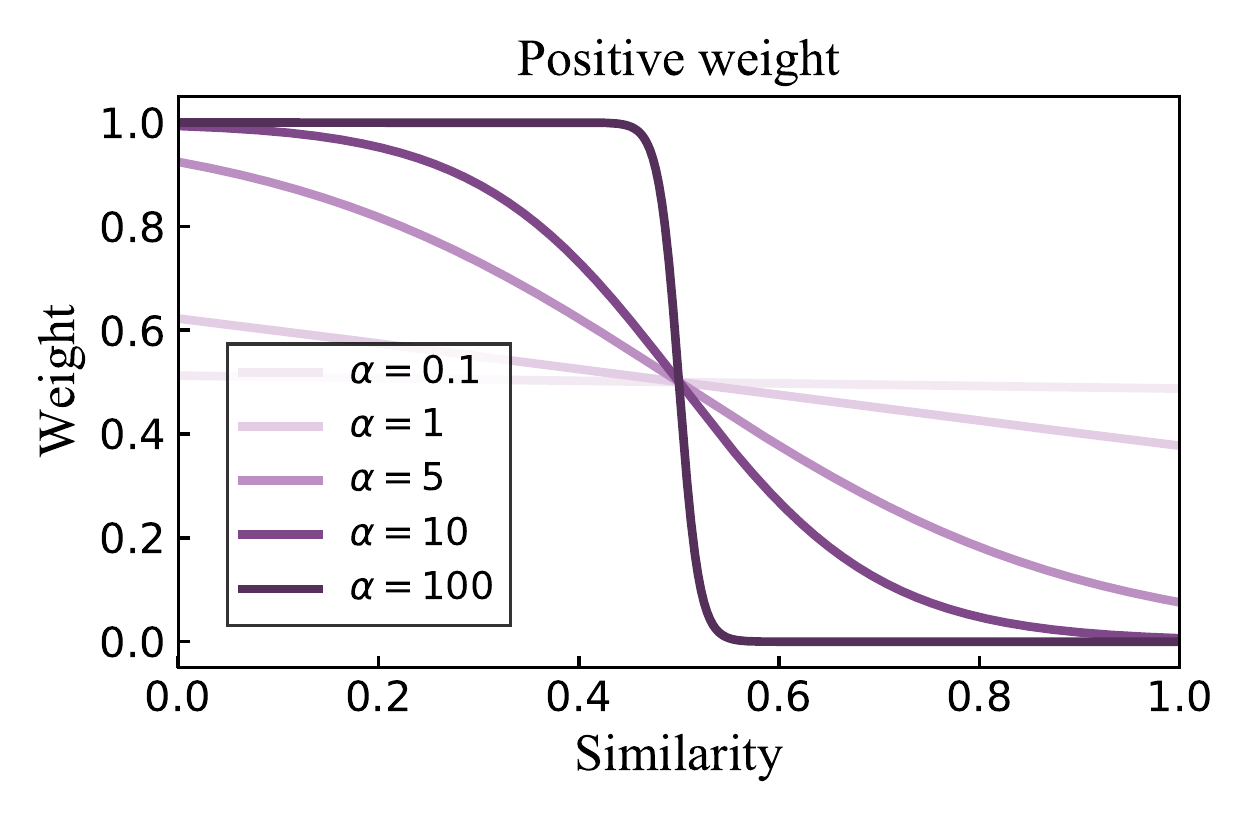}  
      \end{subfigure}
      \begin{subfigure}{0.236\textwidth}
        \centering
        \includegraphics[width=1.0\linewidth, trim=10 10 10 0, clip]{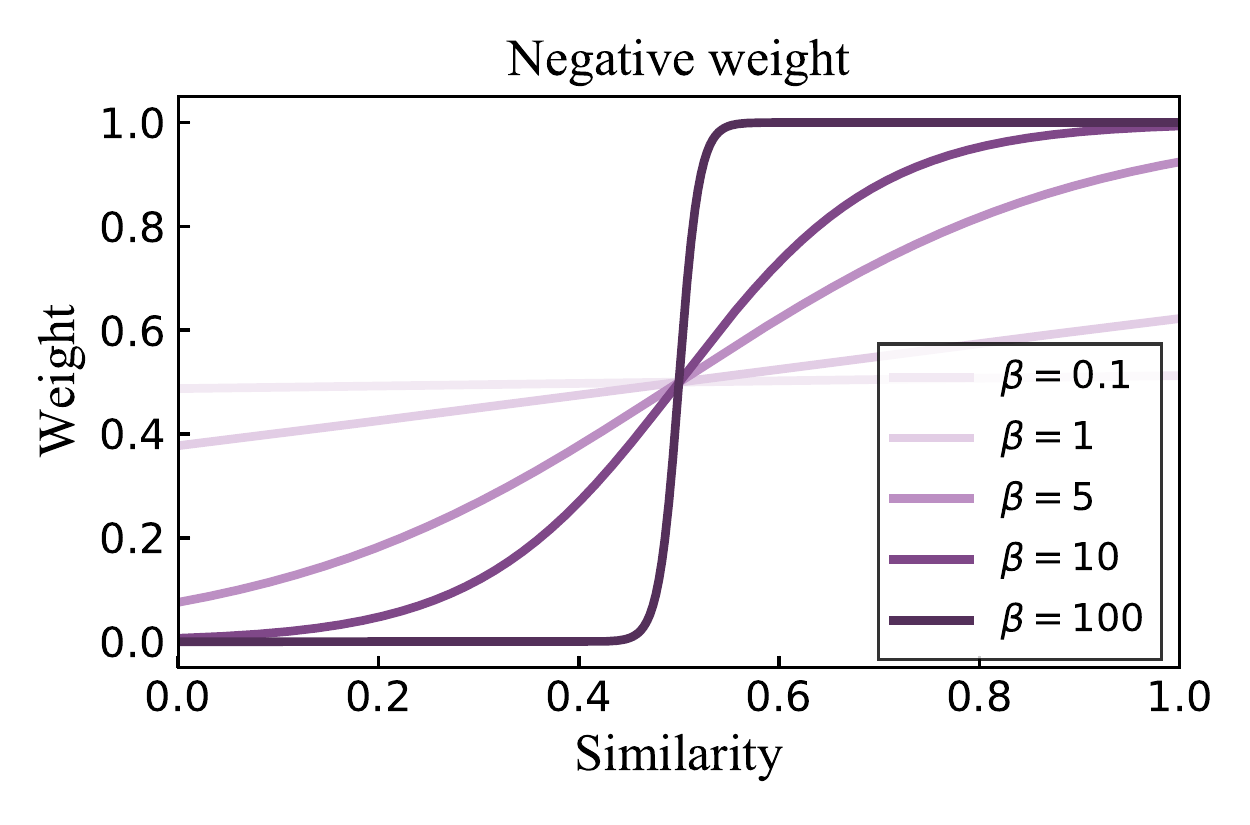}  
      \end{subfigure}
      \vspace{-0.5em}
    \caption{\small \textbf{Positive and negative weight curves by varying scale factors of binominal loss.} A large scale factor induces stronger discriminative ability.}
    \label{fig:binominal}
    \vspace{-1em}
    \end{figure}

\begin{figure}[t]
\begin{subfigure}{0.23\textwidth}
\centering
\includegraphics[width=1.0\linewidth, trim=0 10 0 0, clip]{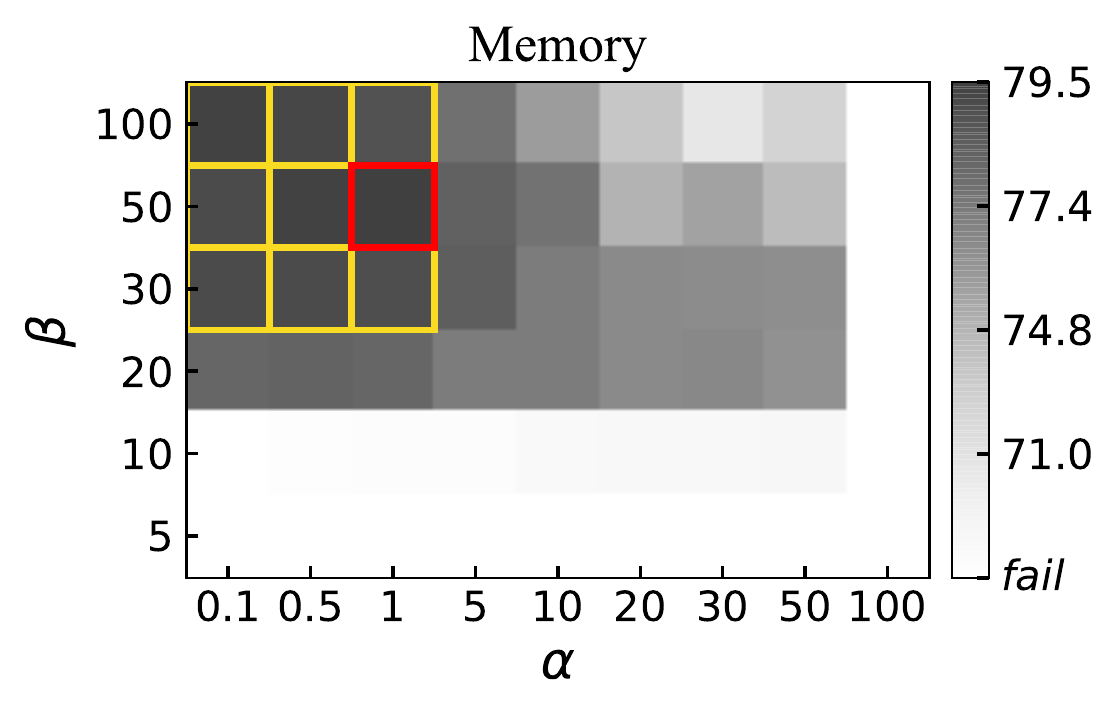}  
\end{subfigure}
\begin{subfigure}{0.23\textwidth}
\centering

\includegraphics[width=1.0\linewidth, trim=0 10 0 0, clip]{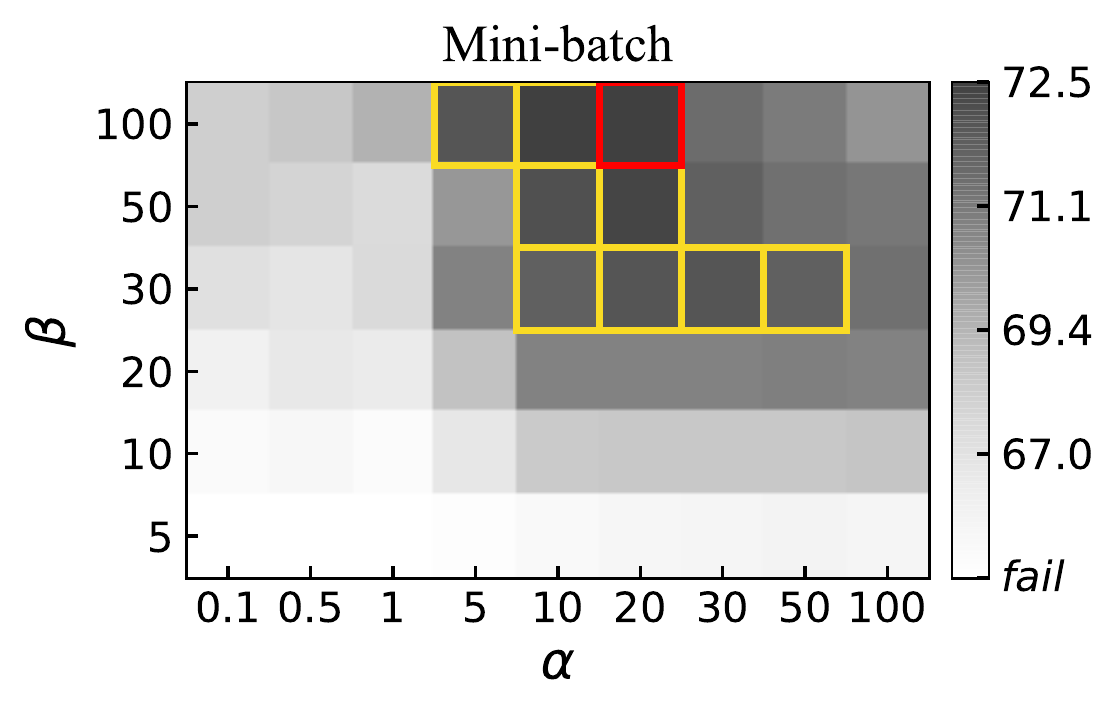} 
\end{subfigure}
\definecolor{myyellow}{RGB}{219, 179, 16}
\vspace{-0.5em}
\caption{\small \textbf{Recall@1 performance with various positive and negative scale factors of binominal loss on SOP.} Deeper grey color indicates better performance. \textcolor{red}{Top 1} and \textcolor{myyellow}{top 9} best settings are annotated by \textcolor{red}{red} and \textcolor{myyellow}{yellow} box. The numerical differences among top 9 models are marginal. See numerical results in SM.}
\label{fig:bino_corr}
\vspace{-1em}
\end{figure}

\subsection{Positive and Negative Weights}
Previous work commonly studies the weights jointly for positive and negative pairs, while the correlation between the two parts has not been well analyzed.
In this section, we first study the correlation of the two parts in  both the memory-based training and mini-batch training, and then further analyze the impact of them separately.

\textbf{Weight correlation.}
With weight assignment formulations in Table \ref{tab:grad}, we can directly adjust the positive and negative weights via scale factors, e.g.,  $\alpha$ for the positive weight and $\beta$ for the negative weight. The weight curves of binominal loss with different values of  $\alpha$  and $\beta$ are presented in Figure \ref{fig:binominal}.
On the positive weights, binominal loss with a small $\alpha$, tends to approach to a constant value which is similar to contrastive loss. When a large $\alpha$ is applied, it obtains a strong hard mining ability by weighting the hard positive pairs (often having a low similarity) more heavily, which is consistent to those losses presented in Figure \ref{fig:loss_grad}. 
On the negative weights,the weight curves have a margin smoothness can be easily adjusted by varying $\beta$. A large $\beta$ weights the hard negatives (with a higher similarity) more significantly. 
Next, we utilize binominal loss to study the correlation of positive and negative weights.

We conduct experiments by using different combinations of  $\alpha$ and $\beta$,  and the results on mini-batch training and memory-based training are presented in Figure \ref{fig:bino_corr}.
First, the top 9 performance in the mini-batch training indicates a strong correlation between the positive and negative weighs, \ie the optimal $\alpha$ is dependent on $\beta$.
Moreover, the binominal loss with a simple parameter grid search, achieves 72.5\% (with $\alpha=20$ and $\beta=100$) in mini-batch training, outperforming default MS loss and InfoNCE loss (Table \ref{tab:compare_mem_loss}).
This further indicates the importance of joint optimization on the positive and negative weights in the mini-batch training.
%
Second, in the memory-based training, the performance can be consistently improved by using a small $\alpha$ and a large $\beta$ in a reasonable range of $\alpha \le 5$ and $\beta \ge 30$, suggesting that the positive and negative weights work more independently. 
Third, the best performance by mini-batch training is achieved with $\alpha=20$, which is significantly larger than that in memory-based training ($\alpha \le 5$). It suggests that the positive pairs have to be weighted carefully in mini-batch training, which is not the case in memory-based training.
%
In summary, our \textbf{Observation One} is that, \emph{the positive weights and negatives weights are correlated strongly in mini-batch training, but empirically work more independently in memory-based training}. This allows us to evaluate them separately, making the study of pair-based methods simpler in memory-based training.


\definecolor{tbgray}{rgb}{0.85, 0.85, 0.85}

\newcolumntype{?}{p{0.86cm}<{\centering}}
\begin{table}[t]
\small
\tablestyle{0.2pt}{1.1}
\begin{tabular}{cV{0.4}????V{0.4}????} \toprule[1pt]
& \multicolumn{4}{cV{0.4}}{s-MoCo} & \multicolumn{4}{c}{mini-batch}  \\
\multirow{-2}{*}{\diagbox{Pos.}{Neg.}} & cont. & bino. & MS & NCE & cont. & bino. & MS & NCE\\ \midrule[0.3pt]
cont. & \cellcolor{tbgray} 79.9 & 77.9 & 75.8 & 77.3 & \cellcolor{tbgray}63.8 & 70.8 & 62.9 & 49.3 \\
bino. & 78.8 & \cellcolor{tbgray}77.8 &  & & 70.3 & \cellcolor{tbgray}67.8 &  &  \\
MS & 78.6 &  & \cellcolor{tbgray} 75.6 & & 67.3 &  & \cellcolor{tbgray}68.4 &  \\
NCE & 77.0 &  &  & \cellcolor{tbgray}77.1 & 67.3 &  &  & \cellcolor{tbgray}72.3 \\ \bottomrule[1.1pt]

\end{tabular}%
\vspace{-0.5em}
\caption{\small \textbf{Performance on SOP with combinations of positive and negative weights from diverse weighting methods.} Mini-batch training enjoys the benefit of more sophisticated weighting techniques, while for memory training, contrastive loss is the best in both positive weighting and negative weighting.}
\label{tab:compare_pos_neg_weight}
\vspace{-1em}
\end{table}%

\textbf{Cross-function correlation.}
We further study the positive and negative weights with more complicated combinations of them over different loss functions.
%
As shown in Table~\ref{tab:compare_pos_neg_weight}, the performances of original losses lie on diagonals.
%
In memory-based training, we notice that the performance can be improved reasonably by using the simpler negative weights of the contrastive loss on the  other well-designed losses (from diagonal to the left column), suggesting that more carefully weighting on the negatives can not lead to performance improvements in memory-based training. Similarly, using the constant positive weights of contrastive loss on the other loss functions even has slightly improvements (from diagonal to the top row). However, in mini-batch training, the well-designed weighting methods, either on positives or on negatives,  often have stronger performance, and using cross-function weights can lead to a significant reduction on performance in some cases (e.g., NCE), due to the mismatch of strong correlated positive and negative weights. 





\begin{figure}[t]
  \centering
  \begin{subfigure}{0.48\textwidth}
  \includegraphics[width=0.49\linewidth, trim=0 0 0 0, clip]{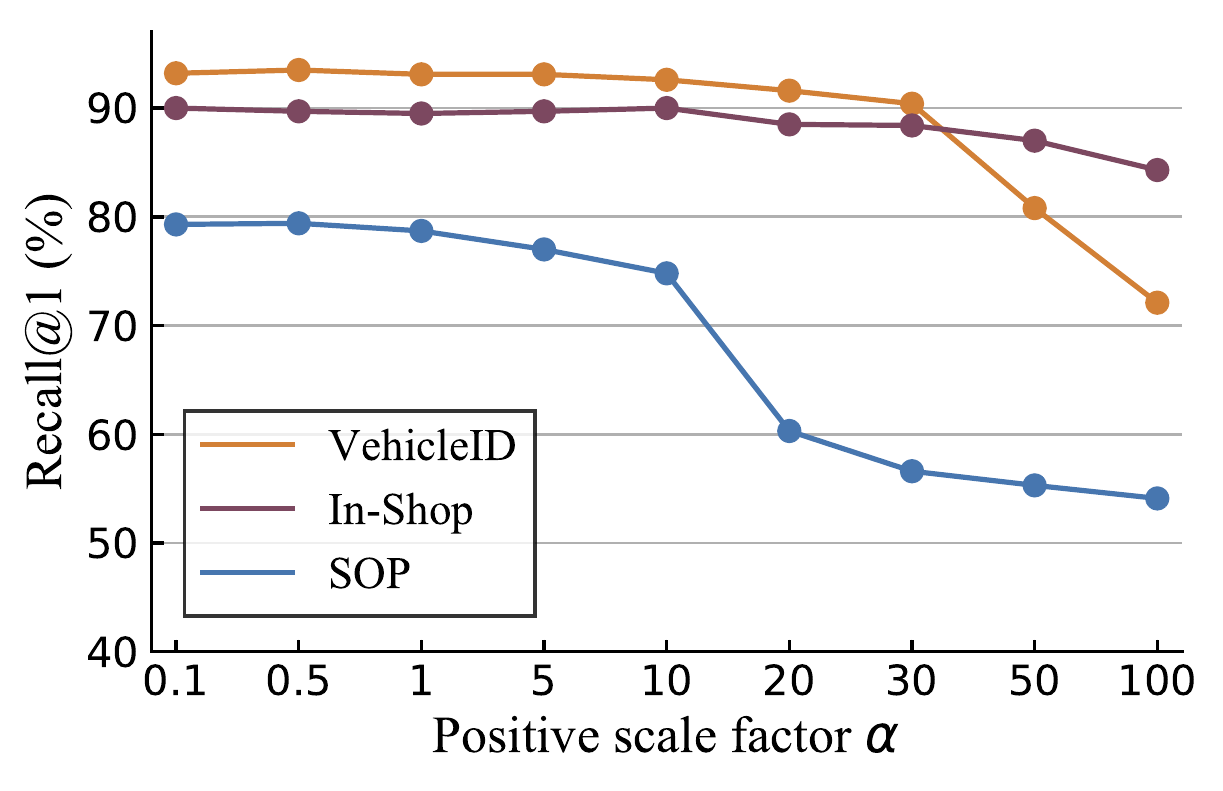}  
  \includegraphics[width=0.49\linewidth, trim=0 0 0 0, clip]{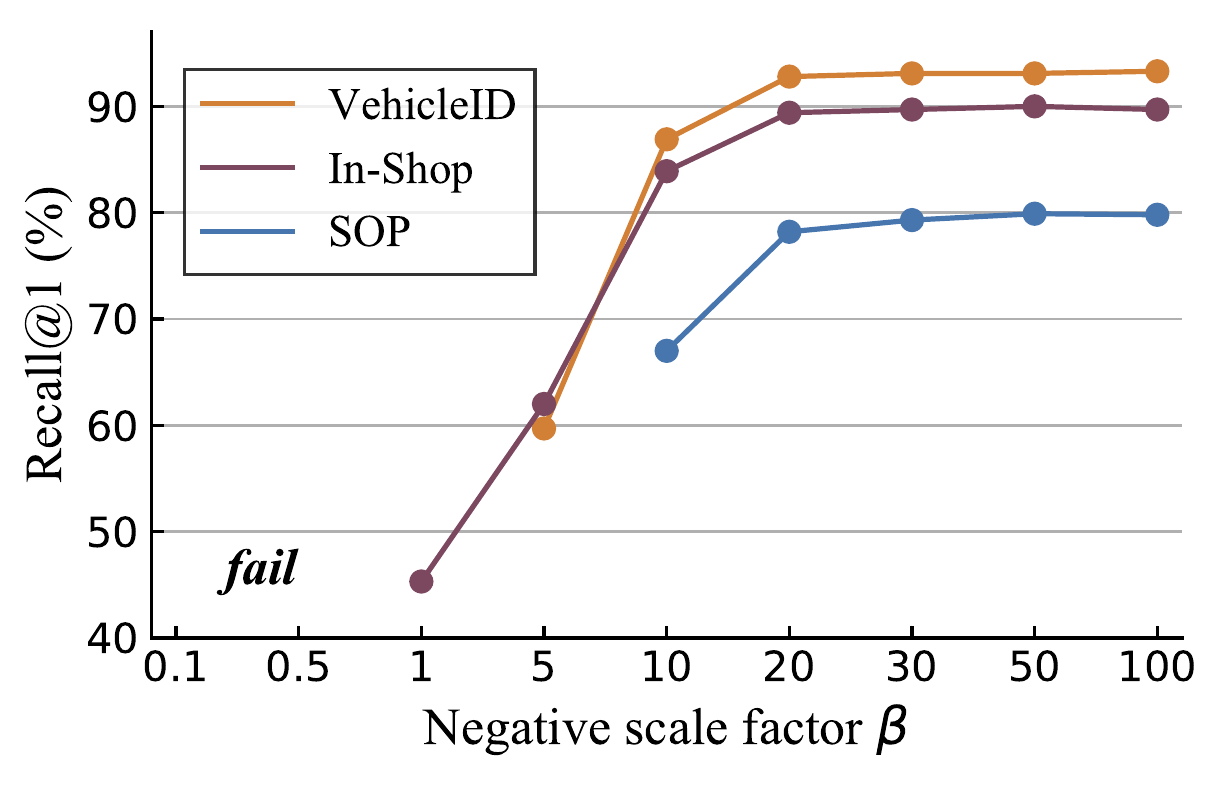}  
  \scriptsize
  \vspace{-1em}
  \caption{Memory}
  \label{fig:pos_vs_neg.a}
  \normalsize
  \end{subfigure}
    \centering
    \begin{subfigure}{0.48\textwidth}
    \includegraphics[width=0.49\linewidth, trim=0 0 0 0, clip]{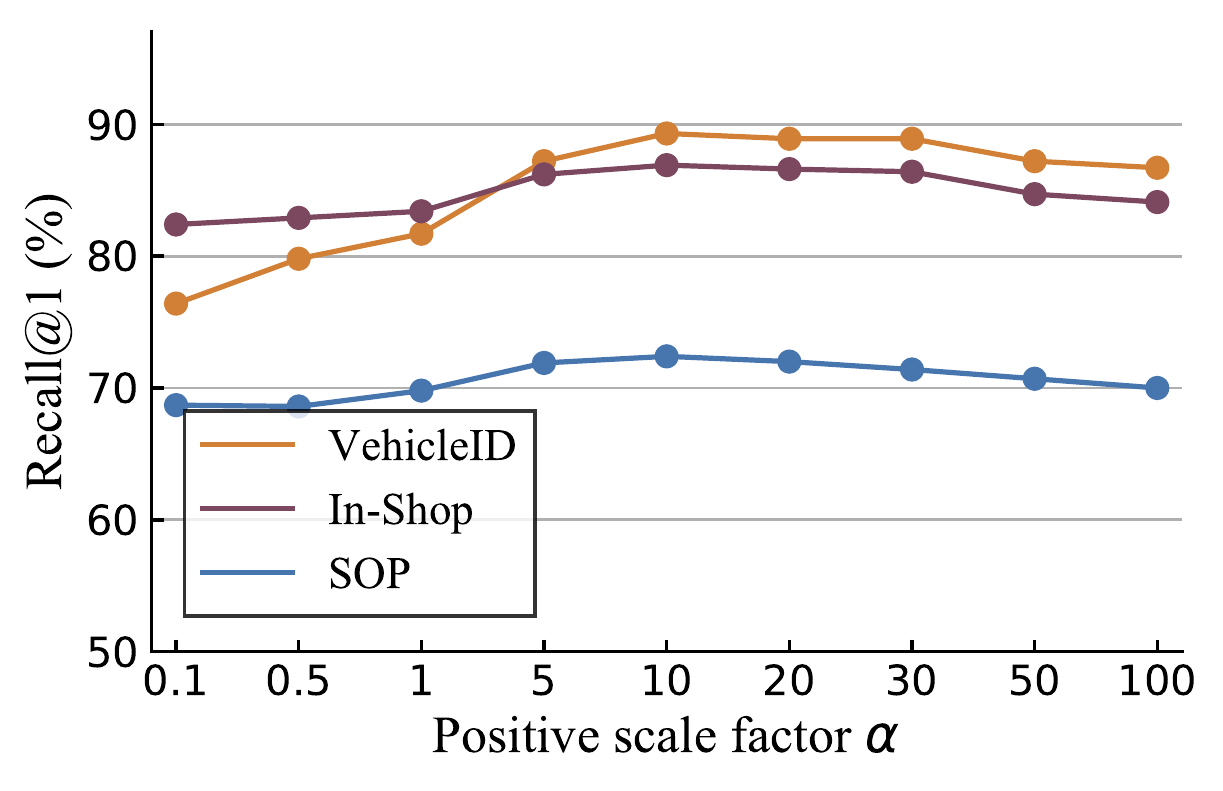}  
    \includegraphics[width=0.49\linewidth, trim=0 0 0 0, clip]{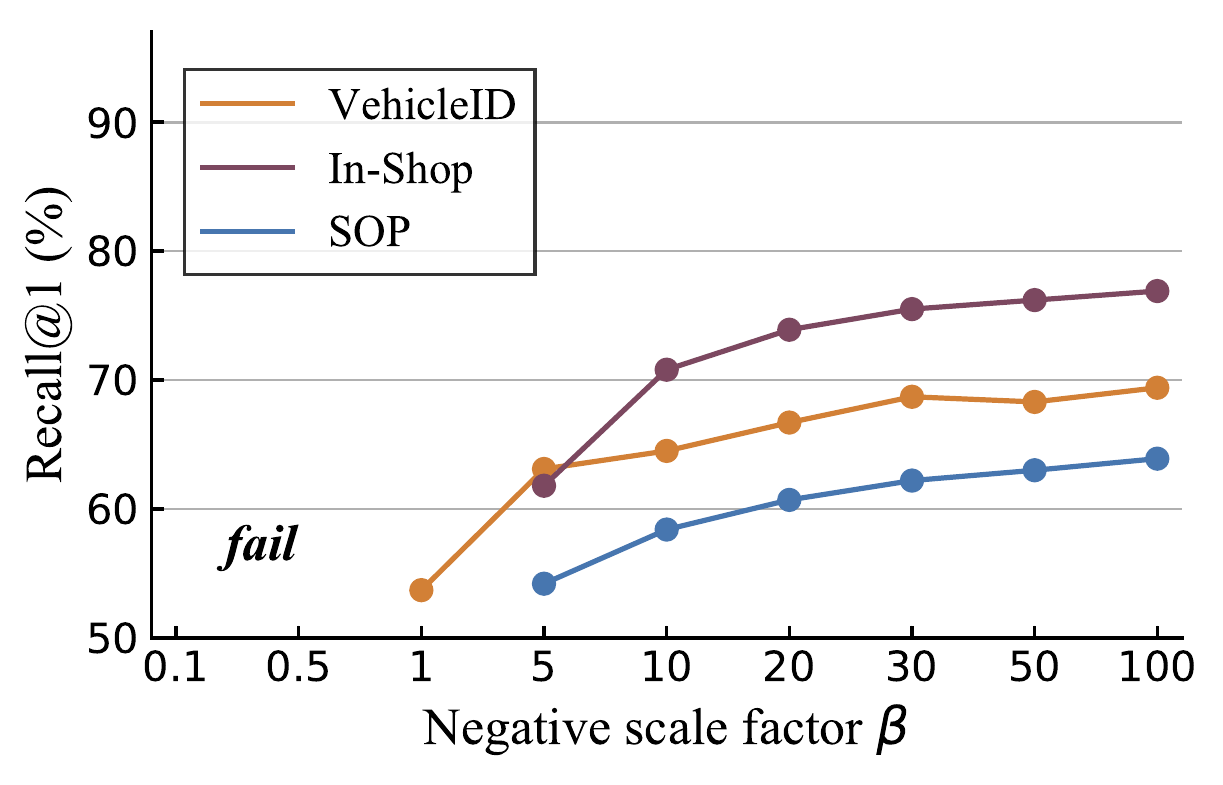}  
    \scriptsize
\vspace{-1em}
  \caption{Mini-batch}
  \label{fig:pos_vs_neg.b}
  \normalsize
  \end{subfigure}
  \vspace{-0.5em}
  \caption{\small \textbf{Performance comparison with various scale factors.} When varying $\alpha$, the negative weights are given by contrastive loss, and vice versa. Data points of collapsed models are not shown.}
  \label{fig:pos_vs_neg}
  \vspace{-1em}
  \end{figure}

\textbf{Key observation.}
The less correlation nature of memory-based training allows us to study the impact of positive and negative weights individually.
Specifically, we vary the positive weights by changing scale factor $\alpha$, and keep the negative weights equal to that of contrastive loss. 
As shown in Figure~\ref{fig:pos_vs_neg.a}, a smaller positive scale, which computes more uniform weights on all positive pairs, consistently obtains higher performance.
For the negative weights, we found that a larger negative scale $\beta$, indicating a sharper weight curve, tends to obtain a stronger model.
However, results of mini-batch training  do not follow this conclusion (Figure~\ref{fig:pos_vs_neg.b}), due to the strong correlation of two parts, which makes the design of weighting schemes more complicated. This comes up with our \textbf{Observation Two}, \textit{with an embedding memory, it is critical to weight the negative pairs, while weighting on the positive ones is trivial, and might be not helpful.}

\begin{figure}[t]
  \begin{subfigure}{0.23\textwidth}
  \centering
  \includegraphics[width=1.0\linewidth, trim=10 10 10 0, clip]{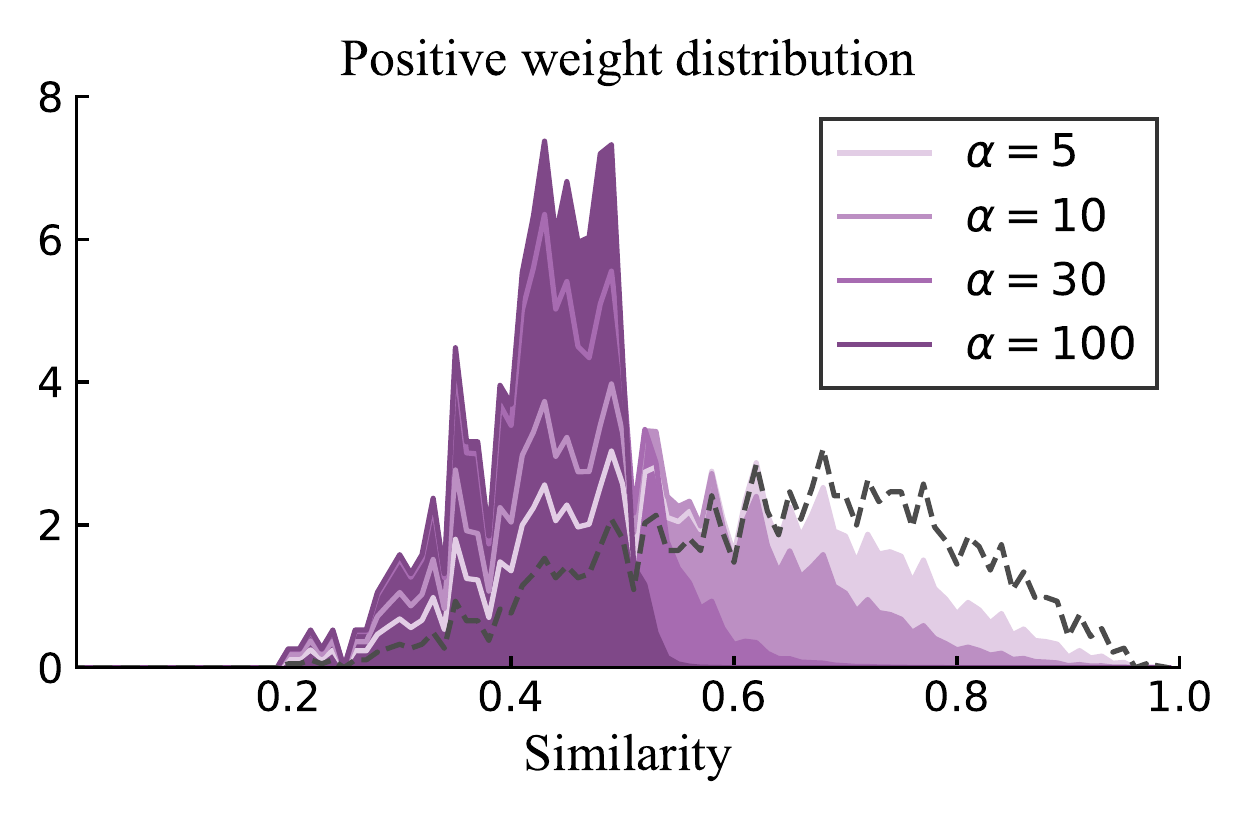}  
  \end{subfigure}
  \begin{subfigure}{0.23\textwidth}
  \centering
  \includegraphics[width=1.0\linewidth, trim=10 10 10 0, clip]{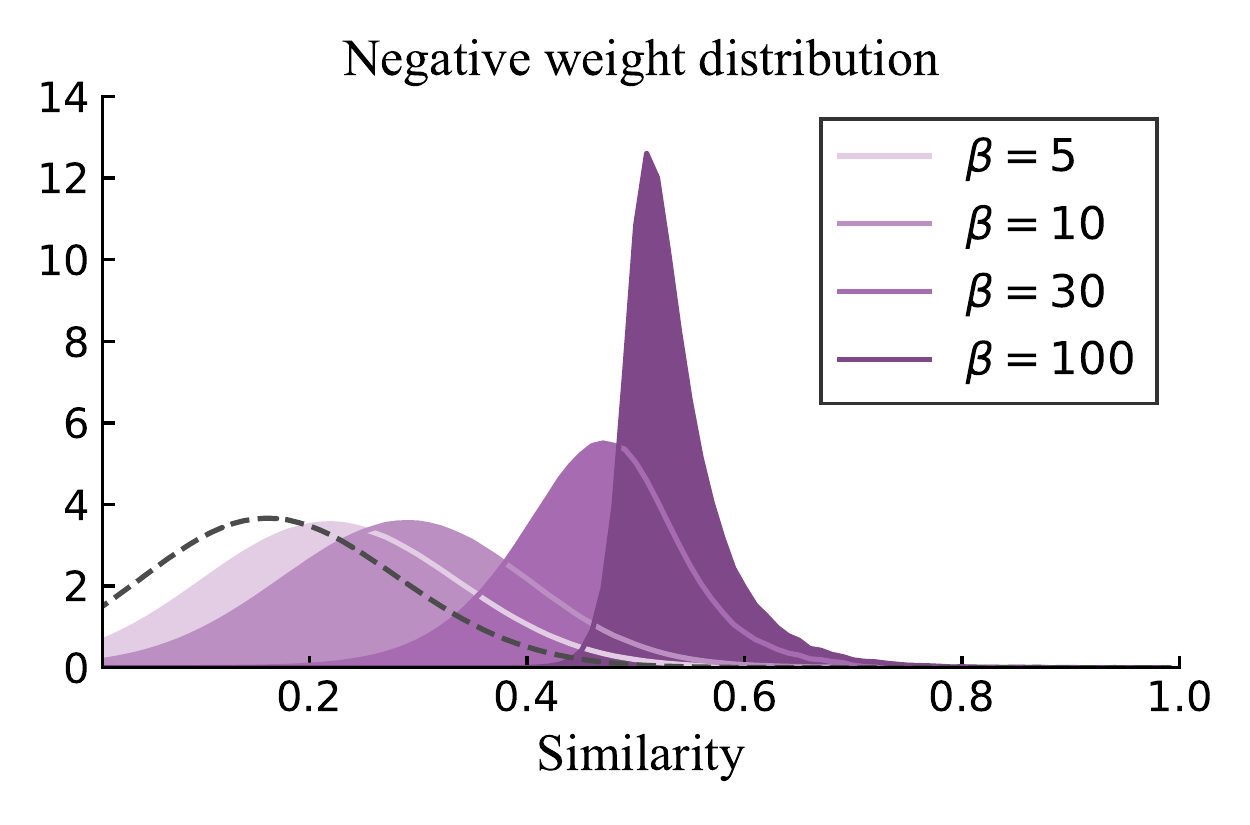}  
  \end{subfigure}
  \vspace{-0.5em}
  \caption{\small \textbf{Positive and negative weight distribution} \wrt similarity of a given batch of pairs (dash line) with various scale factors of binomial loss.}
  \label{fig:bino_dist}
  \vspace{-1em}
  \end{figure}

\textbf{Discussion.} To study the reason of model collapse when a small $\beta$ is used, we visualize the distribution of gradient contribution with respect to similarity in Figure~\ref{fig:bino_dist} Right.
We notice that a large $\beta$ encourages the models to focus on a small number of hard negatives.
On the contrary,  a small $\beta$ will allow for a large number of easy negatives, whose gradients will overwhelm the training, leading to model collapse. 
Whereas positive weights do not have such risk (Figure~\ref{fig:bino_dist} Left).


\subsection {Easy and Hard Negatives}

In this section, we further study the weights on negatives with memory by considering the easy and hard negatives.
Various hard mining strategies have been proposed for DML, \eg marginally discard easy negatives, smoothly transit with a sigmoid-like function and semi-hard mining \cite{facenet}.
We simply consider the negative pairs with a lower similarity than a pre-defined margin as the easy negatives, while  the hard negatives have a similarity higher than the margin. 
In this experiment, we fix the positive weights as a constant value, and study the impact of negative weights by changing the weights of hard negatives and easy negatives separately. 

\begin{figure}[t]
\begin{subfigure}{0.23\textwidth}
\centering
\includegraphics[width=1.0\linewidth, trim=0 0 0 0, clip]{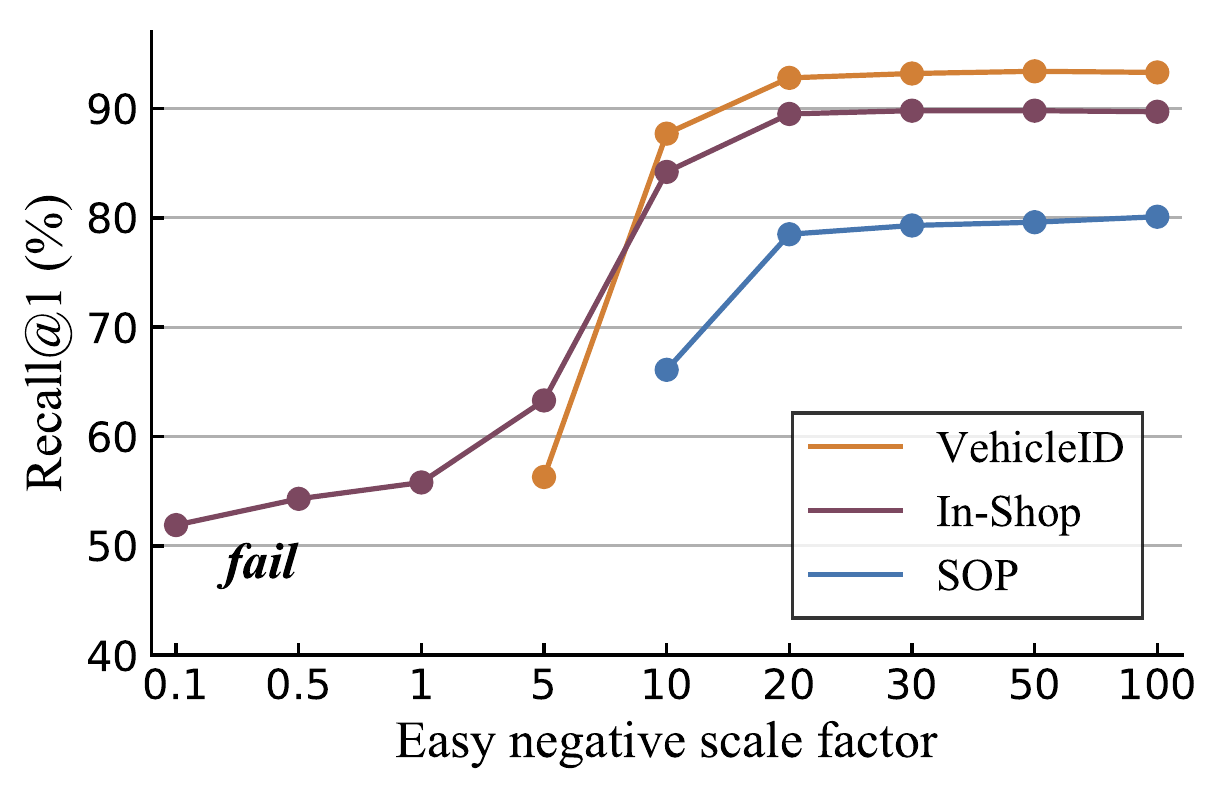}  
\end{subfigure}
\begin{subfigure}{0.23\textwidth}
\centering
\includegraphics[width=1.0\linewidth, trim=0 0 0 0, clip]{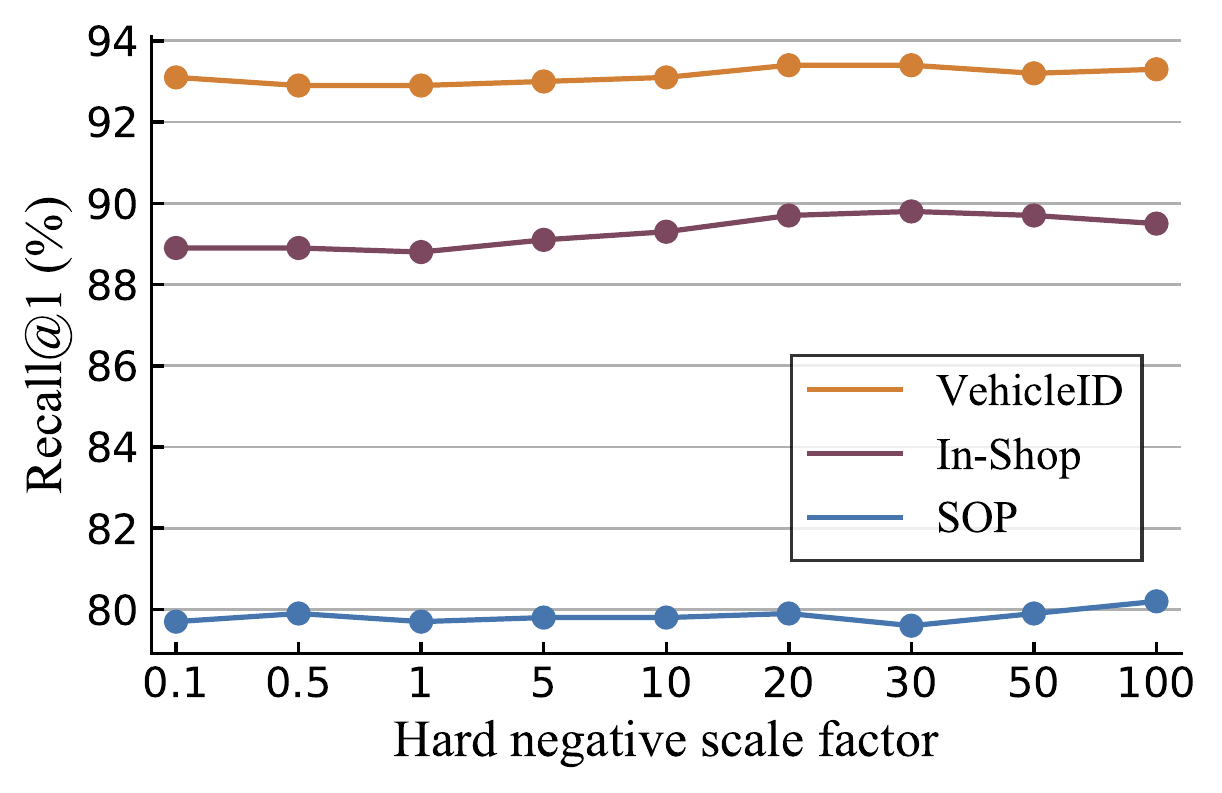}  
\end{subfigure}
\vspace{-0.5em}
\caption{\small \textbf{Comparation of various scale factors on easy and hard negatives.} A large hard negative scale factor can significantly improve the performance, while carefully tuning the scale factor on easy negatives achieves comparable results.}
\label{fig:easy_vs_hard}
\vspace{-1em}
\end{figure}

To evaluate the weighting on easy negatives, we use  the negative weights of binomial loss, and vary the  scale factor ($\beta$) to compute the weights for the easy negatives, while setting the weights of hard negatives to 1. 
As shown in Figure~\ref{fig:easy_vs_hard}, we found that using a large negative scale factor consistently improves the performance, which suggests that assigning small weights to the easy negatives (e.g., reducing the weights closing to 0) can mitigate the side effect from numerous less informative easy negatives.
Similar experiments are conducted on weighting the hard negatives by varying the scale factor, while keeping the weights of easy negatives as 0. 
Interestingly, we found that different hard weighting strategies obtain similar and  consistent performance on all three datasets. 
This comes up with our \textbf{Observation Three} \textit{it is critical to reduce the weights or discard the easy negatives which are less informative but redundant, while carefully weighting the hard negatives is not helpful to memory-based DML.} This is significantly different from the mini-batch training, where the complicated hard weighting approaches (\eg MS loss), and carefully sampling approaches (\eg semi-hard \cite{facenet} and \cite{sampling}), can boost the performance considerably. 



\begin{figure}[t]
\begin{subfigure}{0.155\textwidth}
\centering
\includegraphics[width=1.0\linewidth, trim=0 0 0 0, clip]{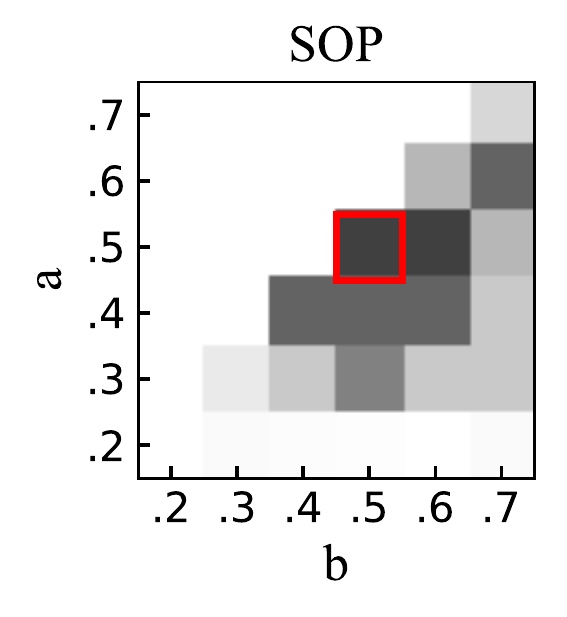}  
\end{subfigure}
\begin{subfigure}{0.139\textwidth}
\centering
\includegraphics[width=1.0\linewidth, trim=0 0 0 0, clip]{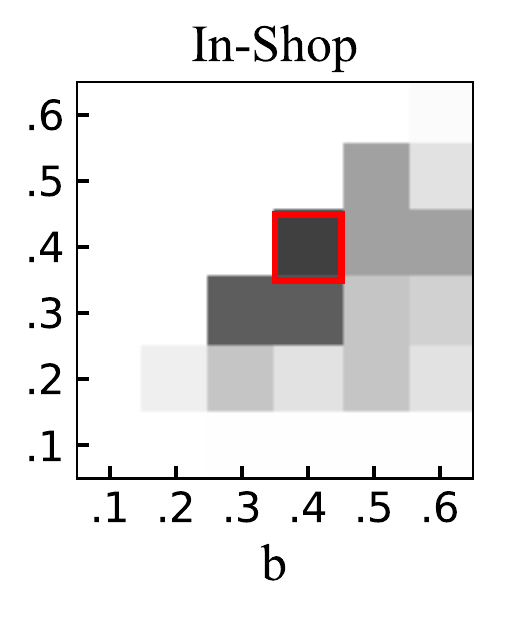}  
\end{subfigure}
\begin{subfigure}{0.174\textwidth}
\centering
\includegraphics[width=1.0\linewidth, trim=0 0 0 0, clip]{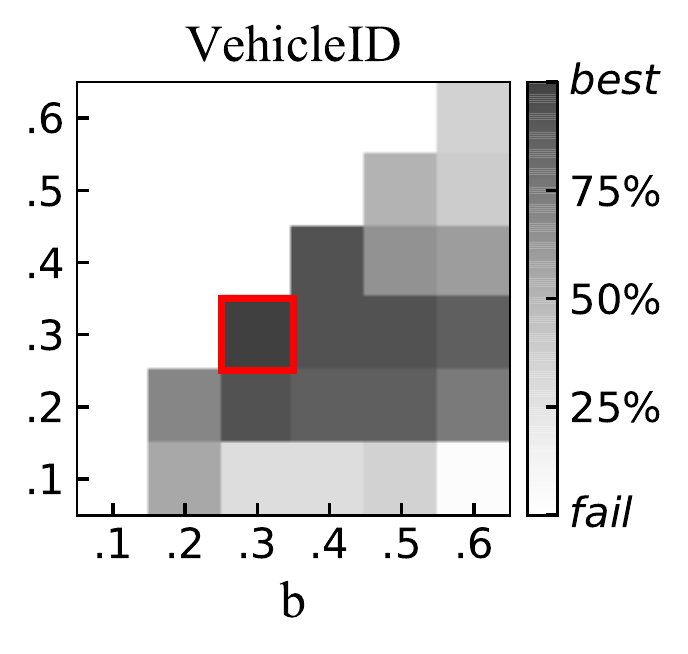}  
\end{subfigure}
\vspace{-0.5em}
\caption{\small \textbf{Relative recall@1 performance of HLL} with various parameter settings. Deeper grey grids indicate better results and white grids indicate invalid parameters or collapsed models. \textcolor{red}{The best settings} are annotated by \textcolor{red}{red boxes}. See SM for numerical results.}
\label{fig:search}
\vspace{-2em}
\end{figure}

\subsection{A Simple Weighting Rule}
\begin{figure*}[t]
\centering
\begin{subfigure}{0.3\textwidth}
\centering
\includegraphics[width=1.0\linewidth, trim=0 0 0 0, clip]{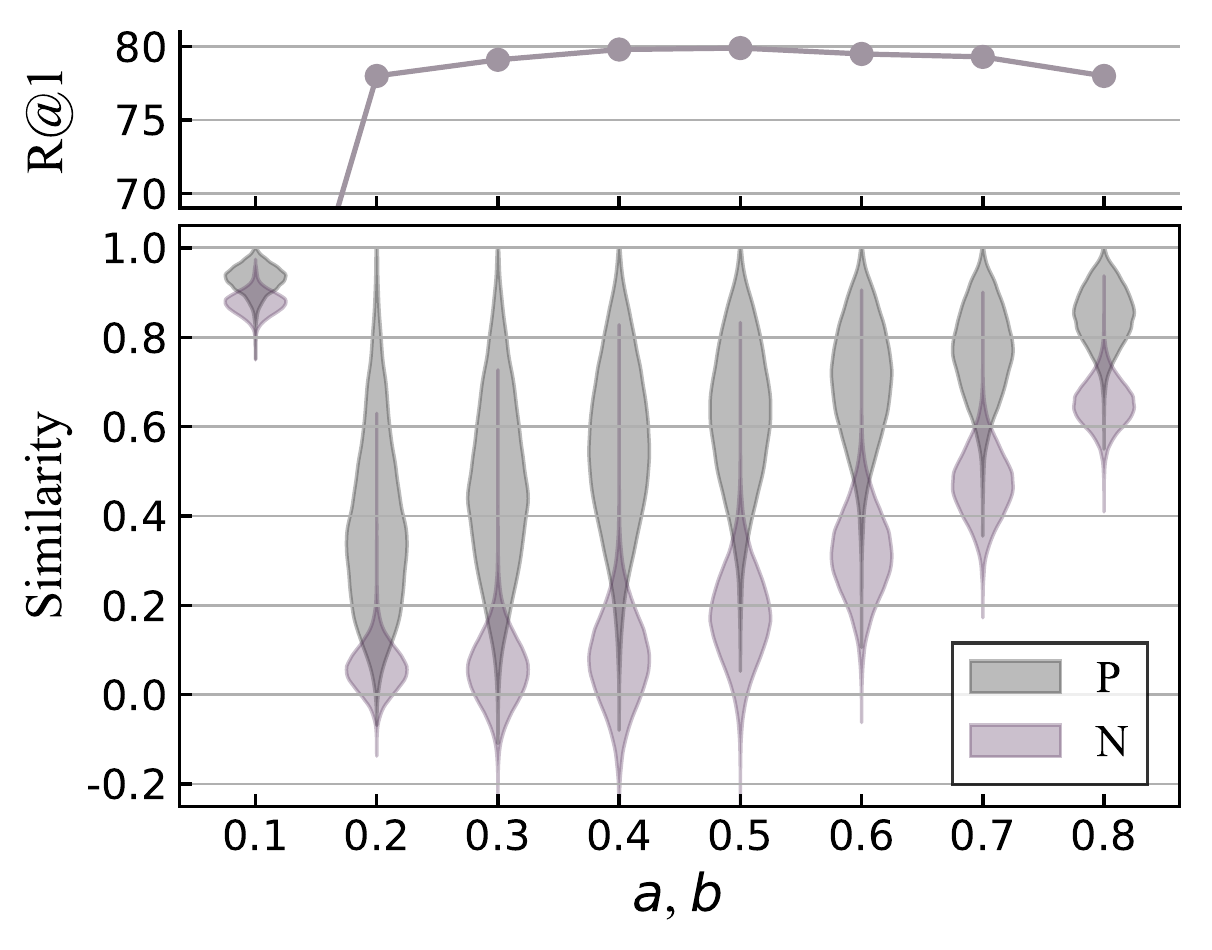}  
\end{subfigure}
\begin{subfigure}{0.3\textwidth}
\centering
\includegraphics[width=1.0\linewidth, trim=0 0 0 0, clip]{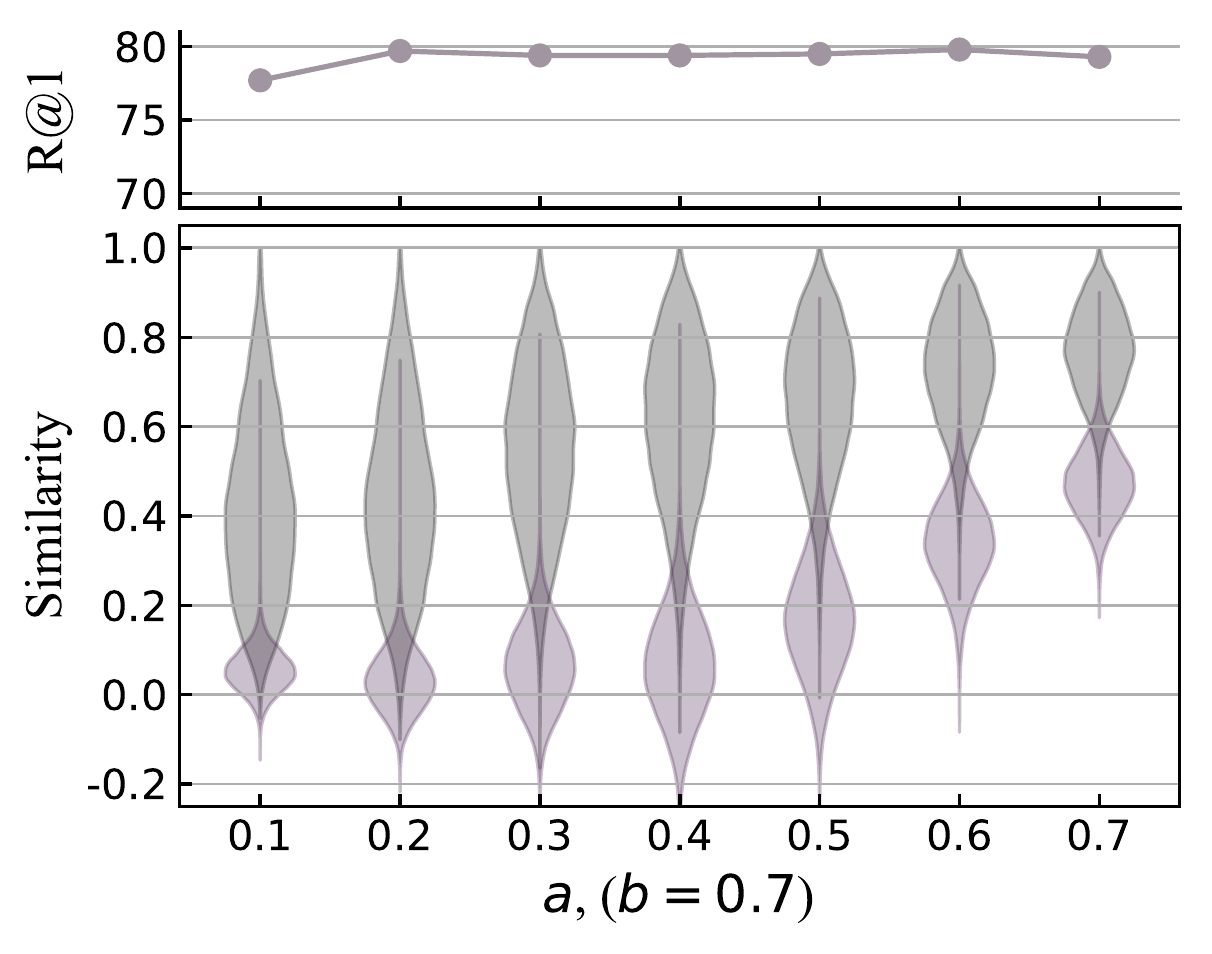}  
\end{subfigure}
\begin{subfigure}{0.3\textwidth}
\centering
\includegraphics[width=1.0\linewidth, trim=0 0 0 0, clip]{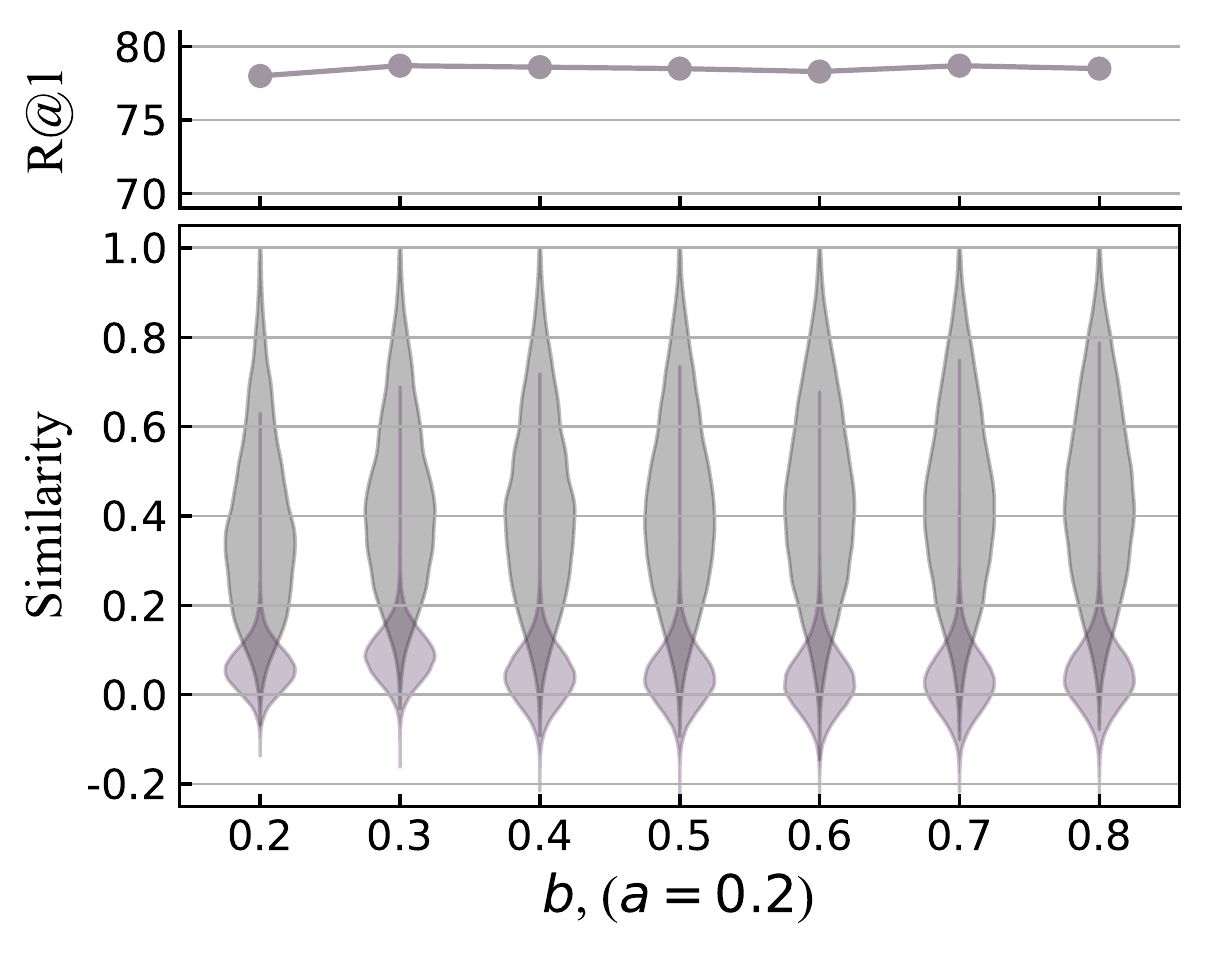}  
\end{subfigure}
\vspace{-0.5em}
\caption{\small Positive and negative distribution with varying $a$ and $b$ of HLL. Recall@1 performances are compared on the top.}
\label{fig:search_dist}
\vspace{-1em}
\end{figure*}

We further utilize a simple hinge-like loss (HLL) function parameterized by two factors to study various negative weighting methods for memory-based DML.
%

Based on our observation two and three, we can streamline the design of weighting scheme, by significantly reducing the search space, which can be simply parameterized using two discrete variables: $a$ and $b$ ($b \ge a$).
Specifically, we weight positive pairs equally, and discard the easy negatives having a similarity lower than $a$, and equally use 1 as the weights of hard negatives which have a similarity higher than $b$. Then  we linearly increase the weights of semi-hard negatives (having a similarity between $a$ and $b$) from 0 to 1.
By adjusting $a$ and $b$, HLL can approximate a wide range of existing pair-based losses with the memory-based DML. For example, we can exactly have a contrative loss with a margin of 0.5 by setting $a = b = 0.5$.

We conduct experiments on three datasets by setting a wide range of the parameters ($a$ and $b$), and demonstrate the full results in Figure~\ref{fig:search}.
The best result on each dataset is highlighted by a red box.
Interestingly, we notice that the best results on three different databases are all achieved with $a=b$, which is equal to using the contrastive loss with a slightly different margin.
This is consistent with the previous observation that carefully weighting the hard negatives using a smoother function such as $a<b$ is not superior to  weighting the hard negatives equally ($a = b$). This further simplifies our weighting scheme, and results in \textbf{a simple weighting rule} for memory-based DML: \textit{search for a similarity threshold that roughly separates the easy and hard negatives, then simply set the weights of all positives and hard negatives equally (e.g., 1), and discard easy negatives (by setting its weights to 0).} 


\textbf{Discussion.}
Since the goal of DML is to discriminate positive and negative pairs, we further study how $a$ and $b$ act on this discriminate ability by observing the distribution of positive and negative similarities.
%
As shown in Figure~\ref{fig:search_dist} Left, with $a=b$, the decision boundary changes with the similarity threshold or margin in a reasonable range from 0.3 to 0.7, but the difference on performance  is marginal. However, the training will be collapsed by using a further small margin (\eg 0.1), which allows for an increasing number of easy negatives. 
On the other hand, using a large margin over 0.8 may not fully span the embedding space, leading to a performance decrease. 
%
Interestingly, the similarity distributions of both positives and negatives can be significantly influenced by the similarity margin, suggesting that such a similarity margin is critical to the learned embedding space.
We further study the cases when $a$ and $b$ are different. As shown in Figure~\ref{fig:search_dist} Middle and Right, the similarity distribution or decision boundary is mainly changed by $a$, while $b$ dose not make a significant impact. This might because $a$ is the key threshold to identify the easy negatives, and the negatives with $>a$ similarities are considered as validated, no matter hard  or semi-hard negatives (determined by $b$). This further confirms our weighting rule and key observations that reducing the side effect of the easy negatives is the most important to memory-based DML.


\section{Experimental Protocol}
\label{sec:exp}
\textbf{Datasets.}
We evaluate our methods on three large-scale datasets on image retrieval: \textbf{SOP} \cite{lifted-structured-loss}, \textbf{In-Shop} \cite{DeepFashion} and \textbf{VehicleID} \cite{liu2016deep}.
The SOP contains 120,053 online product images with 22,634 categories, 59,551 images (11,318 classes) for training, and 60,502 images (11,316 classes) for testing. 
The In-Shop contains 72,712 clothing images of 7,986 classes, and 3,997 classes with 25,882 images as the training set. 
The test set is partitioned into a query set with 14,218 images of 3,985 classes, and a gallery set having 3,985 classes with 12,612 images. 
The VehicleID contains 221,736 surveillance images of 26,267 vehicle categories, where 13,134 classes (110,178 images) are used for training.
, evaluation is conducted
We report results on its three test set which contain 800 classes with 7,332 images, 1600 classes with 12,995
images and 2400 classes with 20,038 images respectively

\textbf{Implementation details.}
We follow the standard experiment settings in \cite{lifted-structured-loss,n-pairs,Opitz2017BIERB,Kim_2018_ECCV_ABE} for fair comparison. 
Specifically, we adopt GoogleNet \cite{inception}, InceptionBN \cite{inceptionbn} and ResNet50 \cite{resnet} as alternative backbone networks. 
We initialize the model with the model parameters pre-trained from ImageNet \cite{ILSVRC15}. 
A fully-connected layer with $l_2$ normalization is added after the global pooling layer. 
The input images are first resized to $256 \times 256$, and then cropped to $224 \times 224$. 
Random crops and random flips are utilized as data augmentation during training. 
For testing, we only use a single center crop to compute the embedding for each instance as \cite{lifted-structured-loss}.
In all experiments, we use an Adam optimizer \cite{kingma:adam} with $5e^{-4}$ weight decay, and reduce the learning rate by 10 at 50\% and 80\% of total iterations.
The optimal initial learning rate is selected from $3e^{-4}$ and $1e^{-4}$, according to different methods, backbone and datasets.
We build a mini-batch with 64 samples containing 4 samples per category.
The speed of momentum update is set as $m=0.999$, and we set the memory size following XBM.

\section{Comparison with State-of-the-art}
\subsection{With XBM}
\label{sec:compare_with_xbm}

\begin{figure}[t]
\begin{subfigure}{0.23\textwidth}
\centering
\includegraphics[width=1.0\linewidth, trim=0 0 0 0, clip]{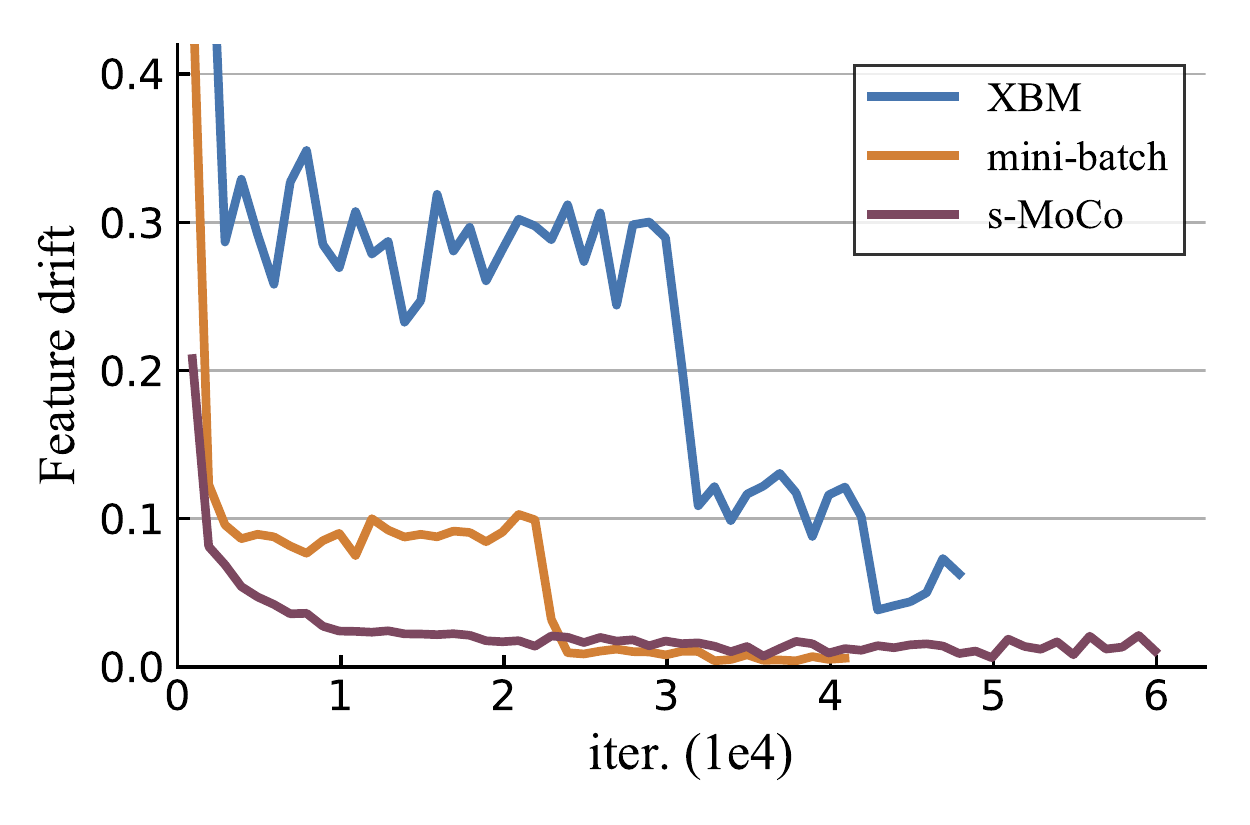}  
\end{subfigure}
\begin{subfigure}{0.23\textwidth}
\centering
\includegraphics[width=1.0\linewidth, trim=0 0 0 0, clip]{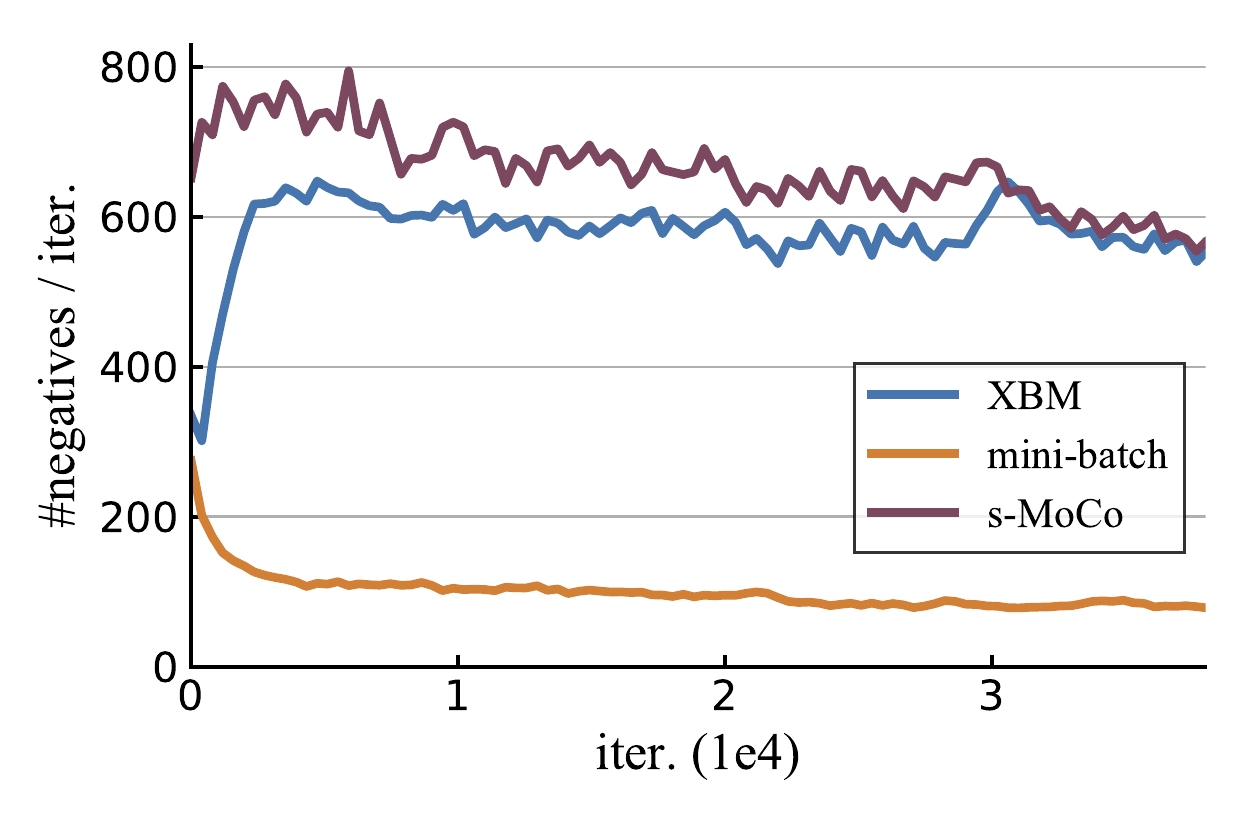}  
\end{subfigure}
\vspace{-1em}
\caption{\small \textbf{Left: Feature drift} of training with mini-batch, XBM and s-MoCo in 1000 steps interval. The feature drift of XBM is significantly higher than mini-batch training. s-MoCo can successfully restrict the feature drift. \textbf{Right: The quantity of hard negatives} per iteration with similarity larger than 0.5.}
\label{fig:compare_xbm}
\vspace{-1em}
\end{figure}


\textbf{Feature drift.} 
As defined in \cite{wang2020cross}, the ``feature drift'' is computed by measuring the feature difference of training samples between different training  iterations. 
%
We compare the 1000 steps feature drift on mini-batch training, XBM and s-MoCo, as shown in Figure~\ref{fig:compare_xbm} Left.
The feature drift of XBM is significantly larger (about 10 times) than that of mini-batch training. 
Obviously, s-MoCo can reduce the feature drift of XBM \textit{significantly}, and is even lower than that of mini-batch training, demonstrating the effectiveness and efficiency of the momentum encoder.  

\textbf{Hard negative mining.}
We further investigate hard mining capability with respect to the feature drift.
We demonstrate the number of  hard negatives per iteration collected by mini-batch method, XBM and s-MoCo  during training. As shown in  Figure~\ref{fig:compare_xbm} Right, s-MoCo mines about 15\% more negatives than XBM, and collects about $\times$7 negatives over mini-batch training.
Therefore, s-MoCo has a stronger hard mining capability over XBM, severing as a strong example of the memory-based DML, which we will compare against the state-of-the-art mini-batch based DML.

\textbf{Momentum.} We study the impact of momentum to the performance by varying its value ($m$) :

\begin{table}[h]
\tablestyle{4pt}{0.9}
\vspace{-1em}
\begin{tabular}{lccccccV{0.4}c} \toprule[1pt]
Momentum ($m$)   & .0 & .5 &.9 & .99 & .999 & .9999 & mini-batch\\ \midrule[0.3pt]
Recall@1(\%) & 77.3 & 77.5 & 78.7 & 78.8 & \bf 79.9 & 78.8 & 63.8\\ \bottomrule[1.1pt]
\end{tabular}%
\vspace{-1em}
\end{table}%

As expected, a small momentum would result in a relatively large feature drift which naturally reduces the performance.  The best performance is achieved at $m = 0.999$, which improves XBM ($m=0$) by 2.6\% Recall@1. But notice that the performance is not changed significantly with different values of momentum, compared to 63.8\% Recall@1 of mini-batch training.



\subsection{Quantitative Results}
\begin{table}[t]
\small
\tablestyle{8.5pt}{0.9}
\begin{tabular}{llcccc} \toprule[1pt]
\multicolumn{2}{l}{Method} & 1 & 10 & 100 & 1000\\ \midrule[0.3pt]
SM \cite{suh2019stochastic} &G$^{\text{512}}$&75.2 &87.5 &93.7 &97.4\\
XBM \cite{wang2020cross} &G$^{\text{512}}$ & 77.4 & 89.6 & 95.4 & 98.4 \\  \midrule[0.3pt]
MS \cite{wang2019multi} &B$^{\text{512}}$& 78.2 &90.5 &96.0 &98.7\\
SoftTriple \cite{qian2019softtriple} &B$^{\text{512}}$& 78.6 & 86.6 & 91.8 & 95.4 \\ 
ProxyGML \cite{zhu2020fewer}& B$^{\text{512}}$ & 78.0 & 90.6 & 96.2 & - \\
Circle \cite{sun2020circle}& B$^{\text{512}}$ & 78.3 & 90.5 & 96.1 & 98.6 \\
XBM \cite{wang2020cross} &B$^{\text{512}}$   & 79.5 & 90.8 & 96.1 & 98.7 \\ \midrule[0.3pt]

FastAP \cite{cakir2019deep} &R$^{\text{128}}$ &73.8 &88.0 &94.9 &98.3\\
MIC \cite{roth2019mic} &R$^{\text{128}}$ & 77.2 &89.4 &95.6& -\\ 
XBM \cite{wang2020cross} & R$^{\text{128}}$   & 80.6 &  91.6 &  96.2 &  98.7 \\ \midrule[0.3pt]\midrule[0.3pt]
s-MoCo &  G$^{\text{512}}$  & 79.9 & 91.1 & 96.2 & 98.7 \\
s-MoCo &  B$^{\text{512}}$  & 80.6 & 91.4 & 96.2 & 98.7 \\
\bf s-MoCo &\bf  R$^{\text{128}}$  &\bf 81.6 &  \bf 92.1 & \bf 96.5 & \bf 98.8 \\ \bottomrule[1.1pt]
\end{tabular}
\vspace{-0.5em}
\caption{\small Recall@$K(\%)$ performance on \textbf{SOP}. Respective backbones are identified with `G' (GoogleNet), `B' (InceptionBN) and `R' (ResNet50). The superscript is embedding size.}
\label{tab:sop}
\vspace{-0.5em}
\end{table}

\begin{table}[t]
\small
\tablestyle{5.5pt}{0.9}

\begin{tabular}{llcccccc} \toprule[1pt]
\multicolumn{2}{l}{Method} & 1 & 10 & 20 & 30 & 40 & 50\\\midrule[0.3pt]
A-BIER  \cite{opitz2018deep} &G$^{\text{512}}$ & 83.1 & 95.1 & 96.9 & 97.5 & 97.8 & 98.0\\	
ABE \cite{Kim_2018_ECCV_ABE}  &G$^{\text{512}}$& 87.3 & 96.7 & 97.9 & 98.2 & 98.5 & 98.7\\
XBM \cite{wang2020cross} &G$^{\text{512}}$ & 89.4 & 97.5 & 98.3 & 98.6 & 98.7 & 98.9  \\\midrule[0.3pt]
HTL \cite{HTL} &B$^{\text{512}}$& 80.9& 94.3& 95.8& 97.2& 97.4& 97.8\\
MS \cite{wang2019multi} &B$^{\text{512}}$& 89.7 &97.9 &98.5 &98.8 &99.1 &99.2\\ 
XBM \cite{wang2020cross} &B$^{\text{512}}$ & 89.9 & 97.6 & 98.4 & 98.6 & 98.8 & 98.9 \\\midrule[0.3pt]
MIC \cite{roth2019mic} &R$^{\text{128}}$ &88.2 &97.0 &- &98.0 &-&98.8 \\
FastAP \cite{cakir2019deep} &R$^{\text{128}}$& 90.9 &97.7 &98.5 &98.8 &98.9 &99.1 \\
XBM \cite{wang2020cross} & R$^{\text{128}}$ &  91.3 &  97.8 &  98.4 &  98.7 &  99.0 &  99.1 \\\midrule[0.3pt]\midrule[0.3pt]
s-MoCo & G$^{\text{512}}$  & 90.3 & 97.5 & 98.4 & 98.6 & \bf 98.9 & 99.0 \\ 
s-MoCo & B$^{\text{512}}$ & 90.7 & 97.8 & 98.5 & 98.7 & \bf 98.9 & 99.1 \\
\bf s-MoCo &\bf R$^{\text{128}}$  & \bf 91.9 & \bf 98.1 & \bf 98.6 & \bf 99.0 & 99.1 & \bf 99.2 \\ \bottomrule[1.1pt]

\end{tabular}
\vspace{-0.5em}

\caption{\small Recall@$K(\%)$ performance on \textbf{In-Shop}.}
\label{tab:inshop}
\vspace{-1em}
\end{table}

\begin{table}[t]
\small
\tablestyle{5.4pt}{0.9}
\begin{tabular}{llcccccc}  \toprule[1pt]
    \multicolumn{1}{l}{\multirow{2}[2]{*}{Method}} 
    && \multicolumn{2}{c}{Small} & \multicolumn{2}{c}{Medium} & \multicolumn{2}{c}{Large} \\
    && 1   & 5   & 1   & 5   & 1   & 5 \\ \midrule[0.3pt]
A-BIER \cite{opitz2018deep}&G$^{\text{512}}$ & 86.3 & 92.7& 83.3& 88.7& 81.9  & 88.7  \\
VANet \cite{Chu_2019_ICCV} &G$^{\text{2048}}$ & 83.3 & 95.9 & 81.1 & 84.7 &77.2 &92.9\\
XBM \cite{wang2020cross} & G$^{\text{512}}$  & 94.0 & 96.3 & 93.2 & 95.4 & 92.5 & 95.5  \\ \midrule[0.3pt]
MS \cite{wang2019multi} &B$^{\text{512}}$& 91.0 & 96.1 & 89.4 & 94.8 & 86.7 & 93.8\\
XBM \cite{wang2020cross} & B$^{\text{512}}$ & 94.6 & 96.9 & 93.4 & 96.0 &93.0 & 96.1  \\\midrule[0.3pt]
MIC \cite{roth2019mic} & R$^{\text{128}}$ & 86.9 &93.4& -& - &82.0 &91.0 \\
FastAP \cite{cakir2019deep} &R$^{\text{128}}$  & 91.9 &96.8 &90.6 &95.9 &87.5 &95.1 \\
XBM \cite{wang2020cross} & R$^{\text{128}}$  & 94.7 & 96.8 & 93.7 & 95.8 &  93.0 & 95.8 \\\midrule[0.3pt]\midrule[0.3pt]
\bf s-MoCo & \bf G$^{\text{512}}$  & \bf 95.4& 97.2& \bf 94.4& 96.2& \bf 94.4 & 96.4 \\ 
s-MoCo & B$^{\text{512}}$ & 95.1 & 96.9& 94.1& 96.2& 93.7 & 96.3  \\
s-MoCo & R$^{\text{128}}$  & 95.3 & \bf 97.6 & 94.5 & \bf 96.7 & 93.9 & \bf 96.7 \\ \bottomrule[1.1pt]

\end{tabular}%
\vspace{-0.5em}
\caption{\small Recall@$K(\%)$ performance on \textbf{VehicleID}.}
\label{tab:vehicleid}
\vspace{-1em}
\end{table}

In this section, we compare s-MoCo with the state-of-the-art methods, in particular the mini-batch based DML. 
For a fair comparison, we report  the results of GoogleNet and InceptionBN with 512 embedding dimension and ResNet50 with 128 dimension.
As discussed previously, the weighting rule for memory-based DML can be significantly simplified, e.g., by just setting a similarity threshold for a dataset. In our experiments, we implement s-MoCo by using the similarity threshold of $\lambda=0.5,0.4,0.3$ on SOP, In-Shop and VehicleID.


Results on the SOP are reported in Table~\ref{tab:sop}, where s-MoCo with B$^{\text{512}}$ can outperform the top pair-based method, Circle loss \cite{sun2020circle} by 78.3\%$\rightarrow$80.6\%  on recall@1,  proxy-based ProxyGML \cite{zhu2020fewer} by 2.6\%, and memory-based XBM \cite{wang2020cross} by 1.1\%.
On the VehicleID with the \emph{large} test set,  s-MoCo with G$^{\text{512}}$  improves recall@1 by 92.5\%$\rightarrow$94.4\%.
We notice that the performance gaps between various backbones are marginal in memory-based methods, but are more significant in mini-batch training. 
On the relatively small In-Shop dataset,  as shown in  Table~\ref{tab:inshop}, we also achieves 0.9\% recall@1 improvement with G$^{\text{512}}$.
Besides, s-MoCo consistently achieves the best performance on other more informative accuracy metrics, \eg R-precision and MAP@R \cite{musgrave2020metric}, as presented in Supplementary Materials (SM). 
See visualization of image retrieval results in SM.

Interestingly, the memory-based methods, including XBM and s-MoCo, cannot comparable with various state-of-the-art mini-batch based loss functions on small-scale datasets: CUB-200-2011 \cite{CUB_200_2011} and Cars-196 \cite{car-196}. 
We conjecture that the memory-based methods enhance class discriminative ability by learning from richer negatives collected from the embedding memory, while how to generalize from train to test is critical in small datasets.
See more discussions in SM.



\section{Conclusion}
In this work, we revisit deep contrastive learning with an embedding memory, and present a new methodology to study pair-based  DML systematically. 
We delve into studying the pair weighting problem by decomposing pair-wise functions, and analyze positive and negative weights separately via weight curves. 
This allows us to observe a number of underlying but insightful facts on pair weighting, and identify a surprisingly simple weighting rule for memory-based DML: performing hard negative mining by setting a similarity margin, as simple as a contractive loss. 
This makes it fundamentally different from existing mini-batch DML, where various pair weighting functions are designed carefully. 
This might give a new direction for DML research, with strong baseline (s-MoCo) provided.



{\small
\bibliographystyle{ieee_fullname}
\bibliography{egbib}
}
\clearpage
\clearpage


\section*{Supplementary material}

# Rethinking Deep Contrastive Learning with Embedding Memory
– Supplementary Material –

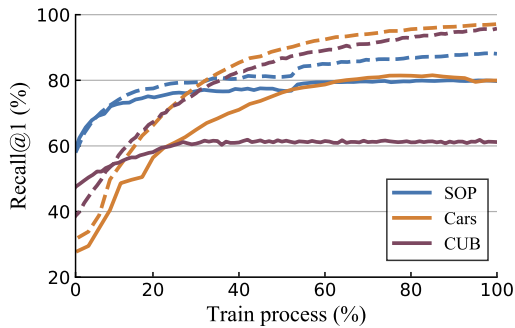

Figure 1: **Recall@1 performance with training process.** A solid line indicates a model is trained on a training set, and a dash line means a model is rained on the testing set (using the same settings). The performance gap between a dash line and its solid line indicates the significance of overfitting on the three datasets.

## 1. More Results on New Evaluation Metrics

In [5], more informative evaluation metrics: Mean Average Precision at R (MAP@R), R-precision (RP), were introduced. We further evaluate our methods with these new metrics in Table 1. Again, s-MoCo consistently outperforms XBM with all evaluation metrics, which verified that reducing the "feature drift" can improve the performance of memory-based DML. Importantly, the results further confirm that the memory-based DML with a simple contrastive loss (s-MoCo) can have large improvements over recent mini-batch DML methods, e.g., MS loss, by using all new metrics in all three benchmarks.

## 2. Results on CUB and Cars-196

We evaluate the performance of s-MoCo on two widely used fine-grained datasets, CUB-200-2011 (CUB) [9] and Cars-196 (Car) [3]. We follow standard train and test protocols. The CUB contains 11,788 birds' images of 200 classes with about 64 images per class. We use 5,864 images of 100 classes for training and the remaining 100 classes for testing. The Cars dataset contains 16,185 images of 196 classes. We use the first 98 categories for training and the remained for testing.

We compare memory-based methods with state-of-the-art mini-batch based approaches in Table 2. We found that

|  |  |  | P@1 | RP | MAP@R |
|---|---|---|---|---|---|
| SOP | MS* | $G^{512}$ | 71.6 | 42.9 | 46.3 |
|  |  | $B^{512}$ | 74.5 | 46.2 | 49.5 |
|  |  | $R^{128}$ | 75.3 | 47.4 | 50.7 |
|  | XBM | $G^{512}$ | 77.1 | 50.1 | 53.2 |
|  |  | $B^{512}$ | 79.4 | 53.1 | 56.1 |
|  |  | $R^{128}$ | 80.8 | 55.2 | 57.9 |
|  | s-MoCo | $G^{512}$ | 79.9 | 53.6 | 56.5 |
|  |  | $B^{512}$ | 80.3 | 54.4 | 57.2 |
|  |  | $R^{128}$ | **81.6** | **55.5** | **58.3** |
| In-Shop | MS* | $G^{512}$ | 85.8 | 60.3 | 63.5 |
|  |  | $B^{512}$ | 88.4 | 64.7 | 67.7 |
|  |  | $R^{128}$ | 88.7 | 64.6 | 67.6 |
|  | XBM | $G^{512}$ | 89.5 | 66.4 | 69.0 |
|  |  | $B^{512}$ | 89.6 | 66.8 | 69.5 |
|  |  | $R^{128}$ | 91.5 | 69.5 | 72.0 |
|  | s-MoCo | $G^{512}$ | 90.2 | 67.9 | 70.4 |
|  |  | $B^{512}$ | 90.7 | 67.9 | 70.5 |
|  |  | $R^{128}$ | **91.6** | **69.8** | **72.2** |
| VehicleID | MS* | $G^{512}$ | 90.5 | 62.5 | 65.6 |
|  |  | $B^{512}$ | 91.8 | 66.0 | 68.6 |
|  |  | $R^{128}$ | 91.4 | 65.3 | 68.1 |
|  | XBM | $G^{512}$ | 92.0 | 65.1 | 67.0 |
|  |  | $B^{512}$ | 93.1 | 69.1 | 70.8 |
|  |  | $R^{128}$ | 93.2 | 69.4 | 71.0 |
|  | s-MoCo | $G^{512}$ | 94.3 | 69.7 | 70.9 |
|  |  | $B^{512}$ | 94.1 | **70.3** | **71.5** |
|  |  | $R^{128}$ | **94.3** | 70.2 | 71.4 |

Table 1: **Evaluation results on new evaluation metrics**: P@1 (precision@1), RP (R-precision) and MAP@R, introduced recently in [5]. * denotes mini-batch training.

the memory-based methods are difficult to outperform the most recent mini-batch approaches on these two *small-scale* datasets, and surprisingly s-MoCo has lower performance than XBM. We conjecture that the embedding memory enables the models with stronger capability which however may make them easier to overfit to a small dataset. A small dataset often has a larger generalization gap between the training and testing sets, and it makes sense that the stronger s-MoCo would have more significant overfitting by using a momentum updated encoder. To further study the over-

| Recall@$K$ (%) | | CUB | | | | | | Cars | | | | | |
|---|---|---|---|---|---|---|---|---|---|---|---|---|---|
| | | 1 | 2 | 4 | 8 | 16 | 32 | 1 | 2 | 4 | 8 | 16 | 32 |
| HDC [8] | $G^{384}$ | 53.6 | 65.7 | 77.0 | 85.6 | 91.5 | 95.5 | 73.7 | 83.2 | 89.5 | 93.8 | 96.7 | 98.4 |
| A-BIER [6] | $G^{512}$ | 57.5 | 68.7 | 78.3 | 86.2 | 91.9 | 95.5 | 82.0 | 89.0 | 93.2 | 96.1 | 97.8 | 98.7 |
| ABE [2] | $G^{512}$ | 60.6 | 71.5 | 79.8 | 87.4 | - | - | 85.2 | 90.5 | 94.0 | 96.1 | - | - |
| XBM [?] | $G^{512}$ | 61.9 | 72.9 | 81.2 | 88.6 | 93.5 | 96.5 | 80.3 | 87.1 | 91.9 | 95.1 | 97.3 | 98.2 |
| ProxyNCA [4] | $B^{64}$ | 49.2 | 61.9 | 67.9 | 72.4 | - | - | 73.2 | 82.4 | 86.4 | 87.8 | - | - |
| HTL [1] | $B^{512}$ | 57.1 | 68.8 | 78.7 | 86.5 | 92.5 | 95.5 | 81.4 | 88.0 | 92.7 | 95.7 | 97.4 | 99.0 |
| MS [10] | $B^{512}$ | 65.7 | **77.0** | **86.3** | **91.2** | **95.0** | **97.3** | 84.1 | 90.4 | 94.0 | 96.5 | 98.0 | 98.9 |
| SoftTriple [7] | $B^{512}$ | 65.4 | 76.4 | 84.5 | 90.4 | - | - | 84.5 | **90.7** | **94.5** | **96.9** | - | - |
| XBM [?] | $B^{512}$ | **65.8** | 75.9 | 84.0 | 89.9 | 94.3 | 97.0 | **82.0** | 88.7 | 93.1 | 96.1 | 97.6 | 98.6 |
| s-MoCo | $G^{512}$ | 59.6 | 71.0 | 80.7 | 87.8 | 92.9 | 96.4 | 79.9 | 86.9 | 91.6 | 94.6 | 97.0 | 98.5 |
| s-MoCo | $B^{512}$ | 59.7 | 70.9 | 81.1 | 88.2 | 93.6 | 96.4 | 81.2 | 88.0 | 92.6 | 95.5 | 97.6 | 98.7 |

Table 2: Recall@$K$(%) performance on **CUB and Cars**.

| | | Memory | | | | | | | | | Mini-batch | | | | | | | | |
|---|---|---|---|---|---|---|---|---|---|---|---|---|---|---|---|---|---|---|---|
| $\beta\backslash\alpha$ | | 0.1 | 0.5 | 1 | 5 | 10 | 20 | 30 | 50 | 100 | 0.1 | 0.5 | 1 | 5 | 10 | 20 | 30 | 50 | 100 |
| 5 | | fail | fail | fail | fail | fail | fail | fail | fail | fail | 57.6 | 57.8 | 58.7 | 61.9 | 63.4 | 64.3 | 64.5 | 64.9 | 64.5 |
| 10 | | 56.5 | 63.0 | 63.9 | 63.4 | 65.6 | 65.9 | 65.9 | 66.6 | fail | 63.1 | 63.9 | 62.7 | 66.5 | 68.5 | 68.6 | 68.6 | 68.6 | 68.8 |
| 20 | | 78.2 | 78.3 | 78.2 | 77.3 | 77.3 | 76.7 | 76.8 | 76.4 | fail | 65.3 | 66.4 | 66.1 | 68.9 | 70.9 | 70.9 | 70.9 | 71.0 | 70.9 |
| 30 | | 79.2 | 79.2 | 79.1 | 78.5 | 77.3 | 76.7 | 76.6 | 76.5 | fail | 67.1 | 66.7 | 67.6 | 70.9 | 71.8 | 72.1 | 72.1 | 71.8 | 71.4 |
| 50 | | 79.2 | 79.4 | **79.5** | 78.4 | 77.7 | 74.8 | 75.6 | 74.4 | fail | 68.3 | 68.0 | 67.5 | 70.3 | 72.2 | 72.4 | 71.8 | 71.4 | 71.2 |
| 100 | | 79.4 | 79.3 | 79.0 | 77.8 | 75.9 | 73.7 | 70.3 | 72.6 | fail | 68.3 | 68.7 | 69.5 | 72.1 | **72.5** | **72.5** | 71.5 | 71.1 | 70.4 |

Table 3: Numerical results of Figure 4.

| | SOP | | | | | | | In-Shop | | | | | | | VehicleID | | | | | |
|---|---|---|---|---|---|---|---|---|---|---|---|---|---|---|---|---|---|---|---|---|
| $a\backslash b$ | 0.2 | 0.3 | 0.4 | 0.5 | 0.6 | 0.7 | | 0.1 | 0.2 | 0.3 | 0.4 | 0.5 | 0.6 | | 0.1 | 0.2 | 0.3 | 0.4 | 0.5 | 0.6 |
| 0.2 | 78.0 | 78.7 | 78.6 | 78.5 | 78.3 | 78.7 | 0.1 | 66.1 | 87.9 | 88.4 | 88.4 | 88.4 | 88.4 | 0.1 | 89.7 | 93.3 | 92.6 | 92.6 | 92.8 | 91.0 |
| 0.3 | | 79.1 | 79.4 | 79.7 | 79.4 | 79.4 | 0.2 | | 89.2 | 89.7 | 89.4 | 89.7 | 89.4 | 0.2 | | 93.6 | 94.0 | 93.9 | 93.9 | 93.7 |
| 0.4 | | | 79.8 | 79.8 | 79.8 | 79.4 | 0.3 | | | 90.2 | 90.2 | 89.7 | 89.6 | 0.3 | | | **94.1** | 94.0 | 94.0 | 93.9 |
| 0.5 | | | | **79.9** | **79.9** | 79.5 | 0.4 | | | | **90.3** | 89.9 | 89.9 | 0.4 | | | | 94.0 | 93.5 | 93.4 |
| 0.6 | | | | | 79.5 | 79.8 | 0.5 | | | | | 89.9 | 89.4 | 0.5 | | | | | 93.2 | 92.9 |
| 0.7 | | | | | | 79.3 | 0.6 | | | | | | 88.7 | 0.6 | | | | | | 92.8 |

Table 4: Numerical results of Figure 8.

fitting issue on different datasets, we train the models (s-MoCo) directly on the test sets of SOP, CUB and Cars, whose training and testing sets have approximately the same numbers of categories and instances. We measure the performance gap between the models trained on the training and testing sets. As shown in Figure 1, it is obvious that our model has significantly larger performance gaps on the small-scale CUB and Cars than the larger SOP dataset. The larger performance gaps suggest more significance of overfitting on the CUB and Cars, resulting in performance reduction on the two datasets.

## 3. Others

The numerical results of Figure 4 and Figure 8 in the main paper are presented in Table 3 and Table 4. We also demonstrate visualization results of the top retrieved images by s-MoCo in Figure 2.



\end{document}